\documentclass[10pt,twocolumn,letterpaper]{article}

\usepackage{multirow}
\usepackage{multicol}
\usepackage{array}
\usepackage{pifont}
\usepackage[table,dvipsnames]{xcolor}

\usepackage[pagenumbers]{cvpr} 

\definecolor{cvprblue}{rgb}{0.21,0.49,0.74}
\usepackage[pagebackref,breaklinks,colorlinks,citecolor=cvprblue]{hyperref}

\def\paperID{1075} 
\def\confName{CVPR}
\def\confYear{2024}

\title{EmbodiedScan: A Holistic Multi-Modal 3D Perception Suite\\Towards Embodied AI}

\author{
Tai Wang$^{1\ast}$,
Xiaohan Mao$^{1,2\ast}$,
Chenming Zhu$^{1,3\ast}$,
Runsen Xu$^{1,4}$,
Ruiyuan Lyu$^{1,5}$,
Peisen Li$^{1,5}$,\\
Xiao Chen$^{1,4}$,
Wenwei Zhang$^1$,
Kai Chen$^1$,
Tianfan Xue$^{1,4}$,
Xihui Liu$^{1,3}$,
Cewu Lu$^2$,\\
Dahua Lin$^{1,4}$,
Jiangmiao Pang$^1$\\
[2mm]
$^1$Shanghai AI Laboratory \quad 
$^2$Shanghai Jiao Tong University \quad
$^3$The University of Hong Kong \quad \\
$^4$The Chinese University of Hong Kong \quad
$^5$Tsinghua University\\
\normalsize{
$^\ast$Equal contribution}
}

\begin{document}

\twocolumn[{%
\renewcommand\twocolumn[1][]{#1}%
\maketitle
\begin{center}
  \centering
  \vspace{-4.5ex}
  \includegraphics[width=1.0\textwidth]{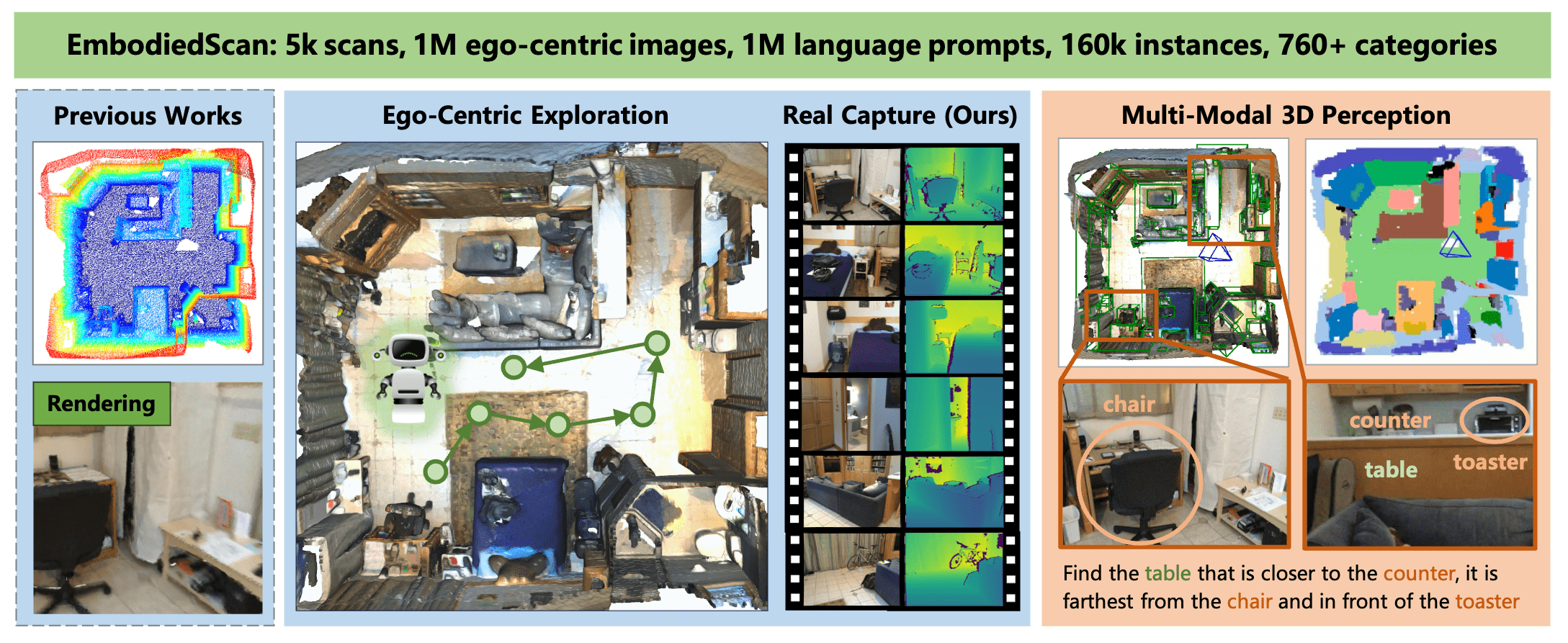}
  \vspace{-4ex}
  \captionof{figure}{EmbodiedScan provides a multi-modal, ego-centric 3D perception dataset with massive real-scanned data and rich annotations for indoor scenes. It benchmarks language-grounded holistic 3D scene understanding capabilities for real-world embodied agents.}
  \label{fig:teaser}
\end{center}%
}]


\begin{abstract}
In the realm of computer vision and robotics, embodied agents are expected to explore their environment and carry out human instructions. 
This necessitates the ability to fully understand 3D scenes given their first-person observations and contextualize them into language for interaction.
However, traditional research focuses more on scene-level input and output setups from a global view.
To address the gap, we introduce EmbodiedScan, a multi-modal, ego-centric 3D perception dataset and benchmark for holistic 3D scene understanding. 
It encompasses over 5k scans encapsulating 1M ego-centric RGB-D views, 1M language prompts, 160k 3D-oriented boxes spanning over 760 categories, some of which partially align with LVIS, and dense semantic occupancy with 80 common categories.
Building upon this database, we introduce a baseline framework named Embodied Perceptron. It is capable of processing an arbitrary number of multi-modal inputs and demonstrates remarkable 3D perception capabilities, both within the two series of benchmarks we set up, \ie, fundamental 3D perception tasks and language-grounded tasks, and in the wild.
Codes, datasets, and benchmarks will be available at \href{https://github.com/OpenRobotLab/EmbodiedScan}{https://github.com/OpenRobotLab/EmbodiedScan}.

\end{abstract}

\begin{figure*}
\vspace{-1ex}
    \begin{minipage}[b]{.66\linewidth}
        \scriptsize
        \centering
        \begin{tabular}{c|ccccccc} 
        \toprule
        Dataset & \#Scans & \#Imgs & \#Objs & \#Cats & \#Prompts & Ego Capture & 3D Annotations\\
        \midrule 
        Replica~\cite{replica} & 35 & - & - & - & - & \ding{55} & \ding{55}\\ 
        NYU v2~\cite{nyuv2} & 464 & 1.4k & 35k & 14 & - & \ding{51} & \ding{55}\\ 
        SUN RGB-D~\cite{sunrgbd} & - & 10k & - & 37 & - &  Mono. & Box \\ 
        ScanNet~\cite{scannet,scannet200} & 1513 & 264k & 36k  & 18 & 52k~\cite{scanrefer} & \ding{51} & Seg., Lang.\\ 
        Matterport3D~\cite{Matterport3D} & 2056 & 194k & 51k & 40 & - & Multi-View & Seg.\\ 
        3RScan~\cite{3rscan} & 1482 & ~363k & - & - & - & \ding{51} & Seg. \\ 
        ArkitScenes~\cite{arkitscenes} & 5047 & 450k & 51k & 17 & - & \ding{51} & Box\\ 
        HyperSim~\cite{hypersim} & 461 & 77k & - & 40+ & - & Mono. \& Syn. & Box\\ 
        \rowcolor{black!12}
        EmbodiedScan & 5185 & 890k & 160k & 762 & 970k & \ding{51} & Box, Occ., Lang.\\
        \bottomrule 
        \end{tabular}
        \captionof{table}{Comparison with other 3D indoor scene datasets. ``Cats" refers to the categories with box annotations for the 3D detection benchmark. EmbodiedScan features more than $10\times$ categories, prompts, and the most diverse annotations. The numbers are still scaling up with further annotations. Mono./Syn./Lang. means Monocular/Synthetic/Language.}
        \label{tab:dataset-table}
    \end{minipage}
    \hspace{2ex}
        \begin{minipage}[b]{.32\linewidth}
        \centering
        \includegraphics[width=6cm]{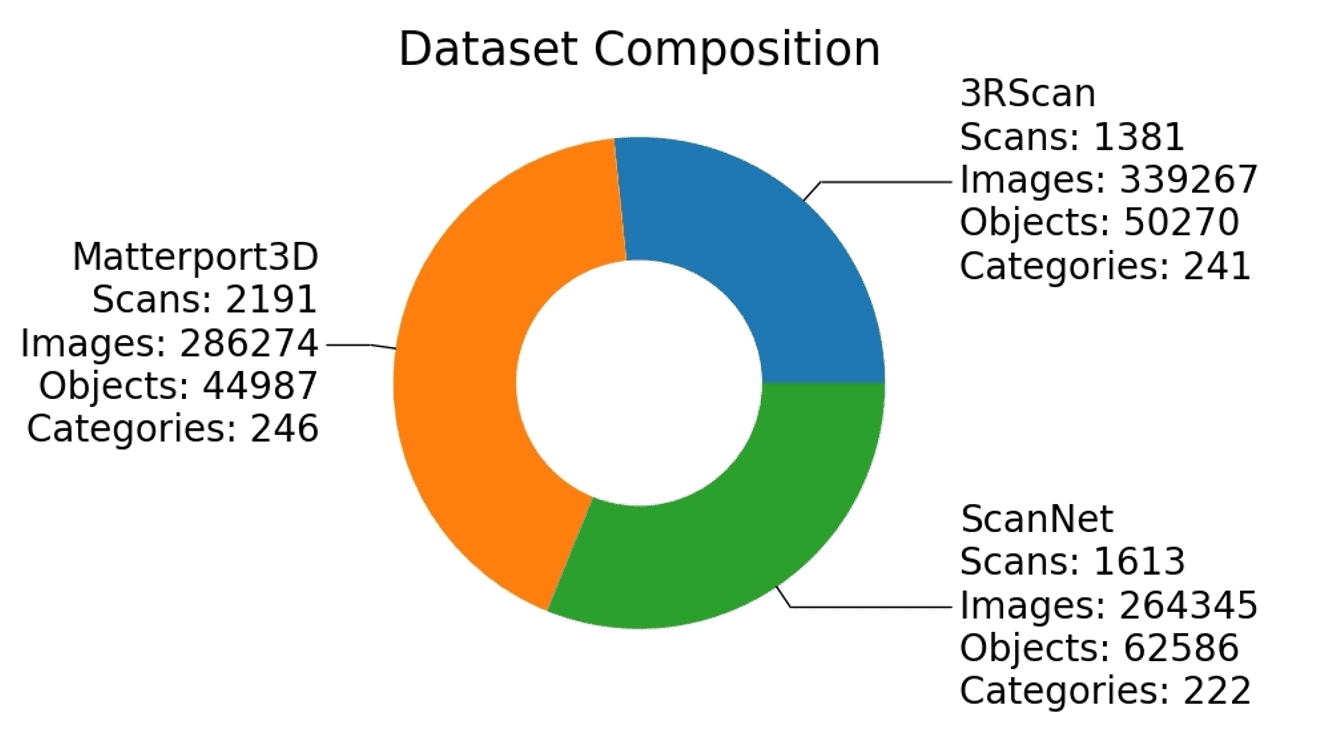}
        \caption{Dataset composition. EmbodiedScan is composed of three data sources and has similar scans, images, objects, and categories in each of them.}
        \label{fig:dataset-composition}
    \end{minipage}
\vspace{-5.5ex}
\end{figure*}

\vspace{-3ex}
\section{Introduction}
\label{sec:introduction}
Consider an embodied agent operating in an indoor environment. It commences its journey devoid of any prior knowledge about the scene, guided only by an initial instruction. As it begins to explore, it recognizes objects in context and acts with goals along with language interaction.
In this process, a commonly needed, fundamental perception capability is to establish a \emph{holistic} 3D scene understanding given \emph{ego-centric} observations. This understanding operates at the scene level, covers both object semantics and scene geometry, and can be grounded in language descriptions.

Nonetheless, subtle but significant discrepancies exist between this expectation and research problems examined within the computer vision community.
Most previous studies have primarily revolved around scene-level input and output problems from a global view~\cite{votenet,fcaf3d,scannet}, \emph{i.e.}, taking reconstructed 3D point clouds or meshes as inputs and predicting 3D object bounding boxes or segmenting point clouds. 
Regarding data, earlier datasets targeting ego-centric RGB-D inputs are either too small~\cite{nyuv2,sunrgbd} or lack comprehensive annotations~\cite{Matterport3D,3rscan} to support the aforementioned research. It is also not feasible to generate such realistic views by rendering from the existing imperfect meshes.
On the other hand, since we cannot trivially obtain the reconstruction of a new environment, models trained with scene-level input are not directly applicable in practice.


To bridge this divide, we introduce a multi-modal, ego-centric 3D perception dataset and benchmark for holistic 3D scene understanding, termed \emph{EmbodiedScan}, aimed at facilitating real-world embodied AI applications (Fig.~\ref{fig:teaser}).
This dataset exploits existing large-scale 3D scene datasets~\cite{scannet,3rscan,Matterport3D} but re-purposes them for continuous scene-level perception from the first-view RGB-D streams.
Unlike previous works that offer only point segmentation labels with limited semantics, we employ a SAM-assisted~\cite{SAM} pipeline to annotate objects with oriented 3D bounding boxes and generate language prompts on top.
Consequently, EmbodiedScan provides more than 5k scans, nearly 1M ego-centric RGB-D images, and multi-modality annotations, covering 3D oriented boxes with more than 160k instances spanning over 760 categories, dense semantic occupancy with 80 common categories, and 1M language descriptions focusing on spatial relationships among objects.

Built upon this dataset, we devise a baseline framework named \emph{Embodied Perceptron} for ego-centric 3D perception.
It accepts RGB-D sequences and texts as inputs and manifests scalability and generalizability to any number of views input with encoders shared across different tasks.
With the encoded 2D and 3D features, we employ dense fusion and isomorphic multi-level fusion across them guided by the perspective projection to produce 3D scene and object representations, which are further processed to decode occupancy and 3D box predictions.
The derived 3D representations can be further integrated with text embeddings for 3D visual grounding, thus supporting language-grounded applications.

We establish two series of benchmarks on EmbodiedScan: 1) fundamental 3D perception benchmarks focusing on traditional tasks, including 3D detection and semantic occupancy prediction under different input settings, and 2) a language-grounded scene understanding benchmark with 3D visual grounding as a preliminary exploration.
Experimental results validate the effectiveness of our baseline model on EmbodiedScan and demonstrate its generalization ability in the wild.
Detailed analysis further underscores the value of EmbodiedScan and highlights the primary challenges posed by this new setup.

\begin{figure*}[t!]
\vspace{-1ex}
     \centering
          \begin{subfigure}[b]{0.49\textwidth}
         \centering
         \includegraphics[width=\textwidth]{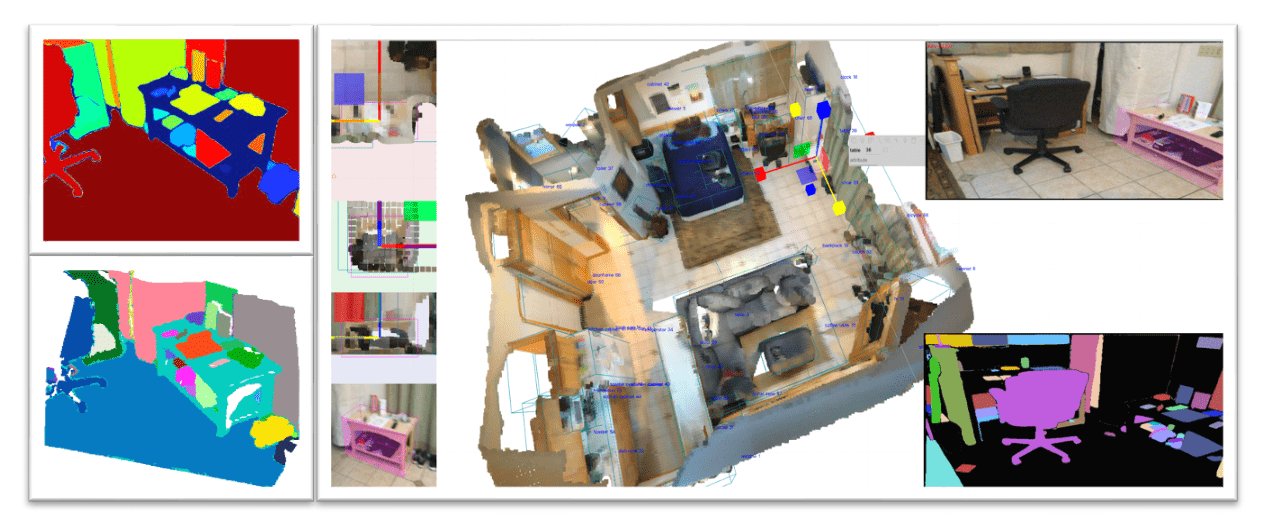}
         \caption{SAM-Assisted Oriented 3D Bounding Boxes Annotation.}
         \label{fig:sam-assisted} 
     \end{subfigure}
     \hfill
     \begin{subfigure}[b]{0.49\textwidth}
        \centering
        \includegraphics[width=\textwidth]{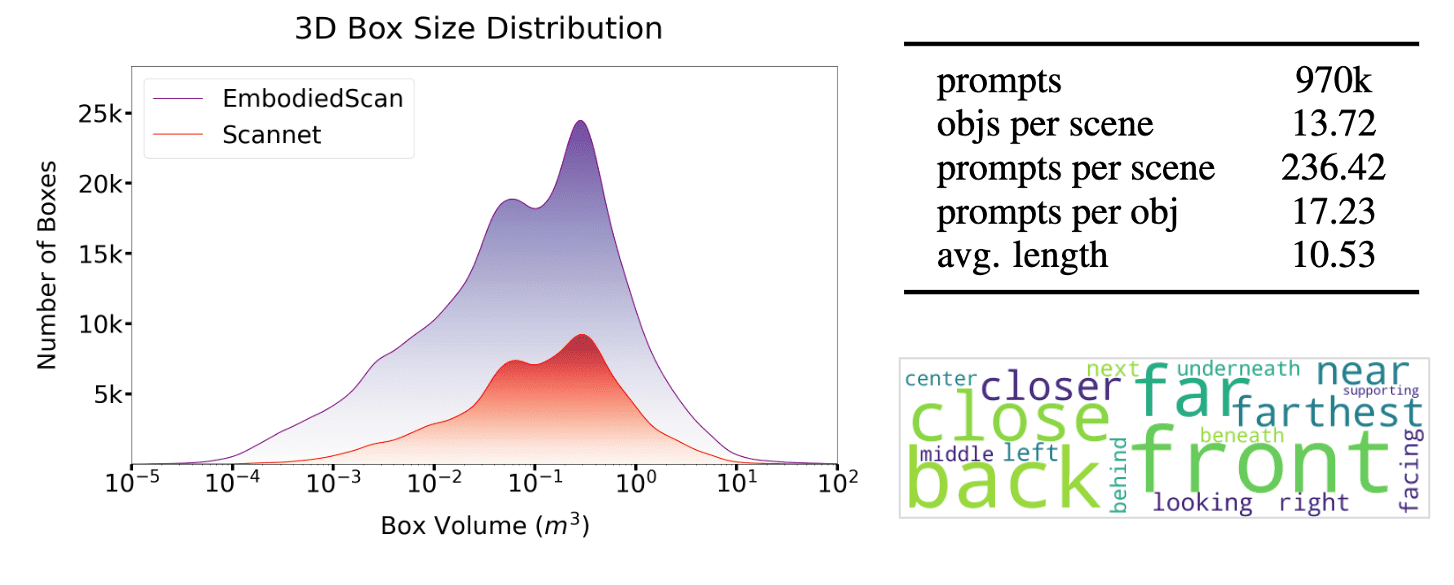}
        \caption{3D Boxes and Language Prompt Statistics.}
        \label{fig:3d-box-prompt-stat} 
     \end{subfigure}
     \\[0.05in]
     \begin{subfigure}[b]{0.49\textwidth}
         \centering
         \includegraphics[width=\textwidth]{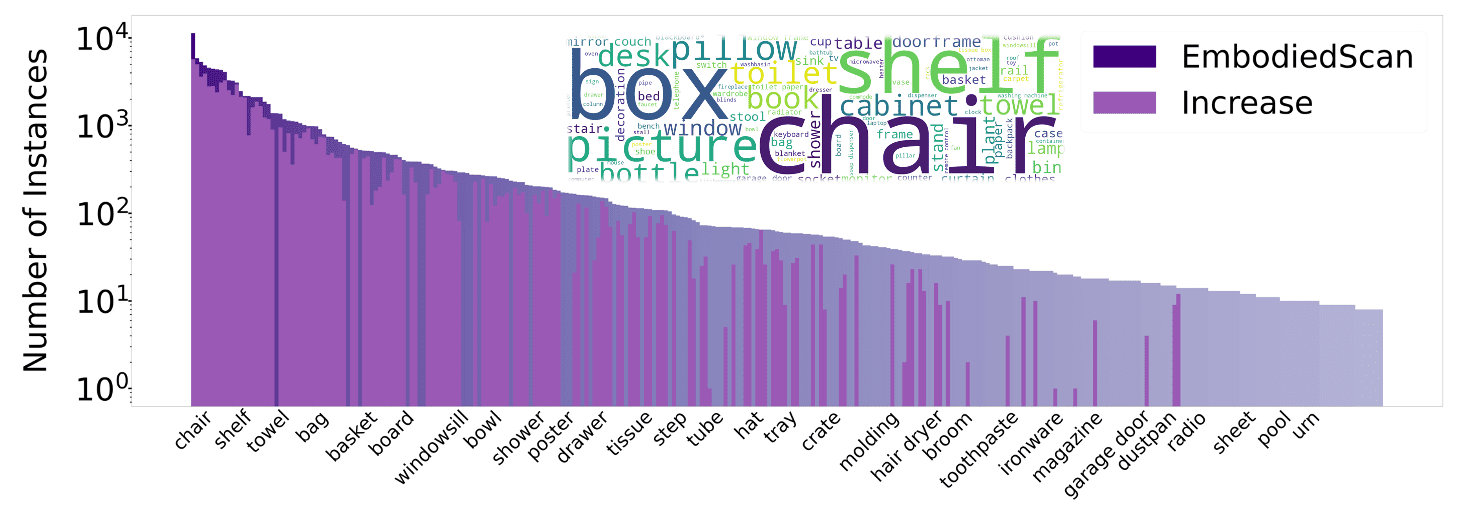}
         \caption{Instance Statistics (Increase w.r.t. ScanNet).}
         \label{fig:instance-stat} 
     \end{subfigure}
     \hfill
     \begin{subfigure}[b]{0.49\textwidth}
         \centering
         \includegraphics[width=\textwidth]{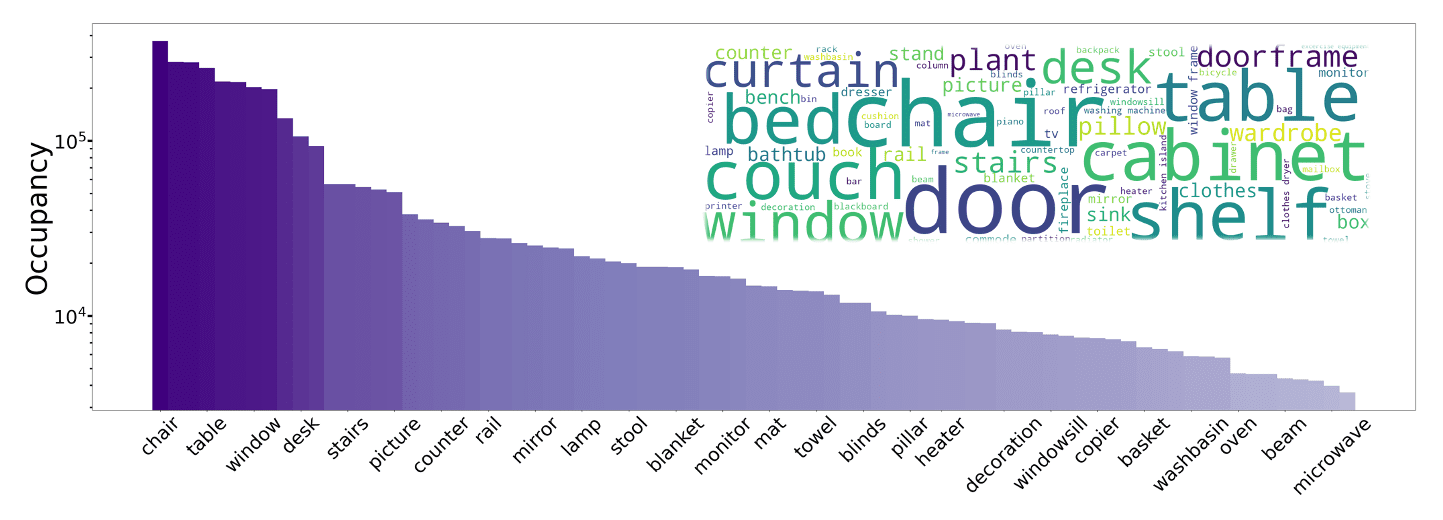}
         \caption{Occupancy Statistics.}
         \label{fig:occ-stat} 
     \end{subfigure}
     \vspace{-0.5ex}
     \caption{EmbodiedScan annotation and statistics. (a) UI for 3D box annotation. We select keyframes and generate their SAM masks with corresponding axis-aligned boxes. With simple clicks, annotators can create 3D boxes for target objects and further adjust them with reference in three orthogonal views and images. (b) Small boxes ($<1m^3$) increase more \& prompt statistics. objs/avg./des. refer to objects/average/descriptions. (c) We show the number of instances per category (300 classes). For categories that exist in ScanNet, we plot the absolute increase and observe a significant improvement. (d) We plot the occupancy distribution for each category and see a different word cloud distribution. These two clouds show different aspects, occupied space vs. number of instances, of this dataset.} 
     \label{fig:dataset-stat}
     \vspace{-1.4em}
\end{figure*}

\section{Related Work}
\label{sec:related}
\noindent\textbf{3D Scene Datasets.}
The development of 3D scene understanding has benefited from large-scale, high-quality datasets like KITTI~\cite{KITTI} and SUN RGB-D~\cite{sunrgbd}. These foundational datasets have paved the way for subsequent larger and more diverse collections targeting indoor~\cite{scannet,Matterport3D,3rscan,hypersim} and driving scenes~\cite{nuScenes,waymo,once,argoverse}. However, compared to autonomous driving datasets, those meant for indoor scenes still lack variety in terms of scenes and object diversity (Tab.~\ref{tab:dataset-table}).
In contrast, EmbodiedScan provides a large amount of multi-modal data with much richer annotations. Furthermore, it differs by placing an emphasis on the ego-centric perspective within its setup, a feature often overlooked in previous works~\cite{scannet,arkitscenes}.


Except for these conventional dataset works, Omni3D~\cite{omni3d} integrates urban and indoor datasets for monocular 3D detection.
Our focus, however, lies in indoor scenes due to their unique challenges but has a larger amount of data and annotations, \emph{e.g.}, more than 3$\times$ images and categories with more than 10 instances.
In addition, we offer a comprehensive exploration of more general problems for ego-centric 3D perception, such as continuous perception and visual grounding.
Other embodied AI datasets like HM3D~\cite{hm3d,hm3d-semantic} and HSSD~\cite{hssd} provide ample interaction opportunities but can suffer from poor transferability to real-world scenarios due to their imperfect meshes or synthetic data. Conversely, EmbodiedScan is based on real-scanned RGB-D images, offering a more realistic playground for model training.


\noindent\textbf{3D Object Detection \& Occupancy Prediction.}
3D detection and occupancy prediction, as fundamental tasks in 3D perception, focus on different aspects of 3D scene understanding. The former focuses on recognizing foreground objects through a sparse and efficient representation - a set of 3D cuboids corresponding to instances of interest - while the latter offers a dense, structured pattern that benefits downstream planning.
The research community has developed solutions ranging from single-modality, such as LiDAR-based~\cite{VoxelNet,SECOND,PointPillars,PointRCNN,votenet,groupfree3d,fcaf3d} or camera-only approaches~\cite{MonoDIS,FCOS3D,smoke,imvoxelnet,pgd,dfm}, to multi-modality techniques~\cite{moca,pointpainting,bevfusion,liang2022bevfusion,imvotenet}. Recently, occupancy as a representation, due to its potential in handling unknown semantics and irregular object shapes, has gained more attention~\cite{monoscene,occnet,tpvformer,occ3d,surroundocc,voxformer}. Given their distinctive focuses, we selected these two tasks to form the fundamental 3D perception track on EmbodiedScan.

Previous works on indoor scenes mainly centered around 3D detection with limitations in object orientations, semantic categories, and input format~\cite{votenet,groupfree3d,fcaf3d}. In practice, a model is expected to perceive the environment during ego-centric exploration, ultimately providing a holistic understanding inclusive of rich semantics, scene geometry, and object poses. To this end, EmbodiedScan and our proposed framework, Embodied Perceptron, provide this necessary data foundation and baseline methodology.

\noindent\textbf{Language-Grounded 3D Scene Understanding.}
Language plays a crucial role in human-computer interaction, heightened by recent advances in Large Language Models (LLMs). Its integration with 3D scene understanding is vital for future embodied agents. Past research first explored 3D visual grounding~\cite{scanrefer,referit3d,butd-detr,mv3d-grounding} and established new benchmarks including 3D dense captioning~\cite{scan2cap,Unit3D}, open-vocabulary 3D segmentation~\cite{openmask3d,openscene,PLA} and detection~\cite{3detic,object2scene}. This paper focuses on 3D visual grounding first, with plans to expand language annotations and benchmarks in the future. Our visual grounding benchmark aligns with the multi-view setting of the basic 3D perception track, taking multiple ego-centric RGB-D images as input, and includes tenfold more complex prompts in our challenging dataset.

\section{Dataset}
\label{sec:dataset}
This section presents the dataset construction, including data processing and annotation, and shows the statistics.

\subsection{Data Collection \& Processing}
\noindent\textbf{Ego-Centric Sensor Data Collection.}
Considering there have been readily available 3D indoor scene scans from existing datasets, we start with integrating those providing ego-centric RGB-D captures with corresponding camera poses. Given the compatibility of ScanNet~\cite{scannet}, 3RScan~\cite{3rscan}, and Matterport3D~\cite{Matterport3D}, we select the high-quality part with necessary annotations to form the initial version of EmbodiedScan (Fig.~\ref{fig:dataset-composition}).
ARKitScenes~\cite{arkitscenes}, possessing different data organization, depth sensors, and annotations, is considered for future inclusion.


\noindent\textbf{Frame Selection \& Scene Division.}
Although these datasets all have RGB-D data, the data format, sampling frequency, and relationships among viewpoints are different. See more details in the appendix.
We first unified the format into a general multi-view case to fit Matterport3D by adding randomness when loading images but maintaining sequential continuity for ScanNet and 3RScan during inference. Our model can thus handle both temporal and randomly captured multi-view images. Additionally, we divided building-scale scenes of Matterport3D into regions based on official annotation, selecting corresponding images with depth points falling into the region. As for different sampling rates of images in ScanNet and 3RScan videos, we sample one keyframe per 10 frames for ScanNet and keep all the images for 3RScan. The uniform sampling is generally in line with the actual situation.

\noindent\textbf{Global Coordinate System.}
A global coordinate system is necessary to aggregate multi-view observations and serve as a reference for outputs. We follow the convention of ScanNet, deriving a system with the origin around the center of the scene, the horizontal plane lying on the floor and axes aligning with walls~\cite{votenet}. This post-processing harmonizes the data distribution, slightly improving performance on benchmarks. Practical applications may not have such a prior global system or vary according to observations, posing another interesting problem for future exploration.


\subsection{Annotation}
We provide three types of annotations - 3D bounding boxes, semantic occupancy, and language descriptions - each serving to enrich different aspects of scene understanding.

\noindent\textbf{3D Bounding Boxes.}
Following standard definitions~\cite{arkitscenes,omni3d}, a cuboid is defined by its 3D center, size, and ZXY Euler angle orientation. We used the Segment Anything Model (SAM)~\cite{SAM} and a customized annotation tool based on \cite{sustech} (Fig.~\ref{fig:sam-assisted}) to address limitations in existing 3D box annotations, \emph{i.e.}, lack of orientation and small object annotations. It supports the conventional functionality of annotating 3D boxes with orientation in three orthographic views. Furthermore, we sample several keyframes with clear imaging according to the camera pose changes and ensure they cover non-overlap regions and most objects to generate SAM masks and axis-aligned boxes for further adjustment. We work with an annotation team and check the quality of all the labels in the end. Each scene takes around 10-30 minutes to annotate, varying with the scene complexity.

\noindent\textbf{Semantic Occupancy.}
Semantic occupancy necessitates accurate boundaries across semantic regions without considering object pose or recalling all the objects, so the original point cloud segmentation annotations were more suitable to be used for deriving occupancy ground truth. For each voxel, we assigned the category with the most points as the semantic label for that cell. A compromise between perception granularity and computational efficiency resulted in $40\times40\times16$ occupancy maps in the perception range $[-3.2m\sim 3.2m, -3.2m\sim 3.2m, -0.78m\sim 1.78m]$ along the X-Y (horizontal) plane and Z (vertical) axis.


\noindent\textbf{Language Descriptions.}
Given updated 3D bounding boxes annotated with orientations, we derive the language prompts that describe the spatial relationships among objects following SR3D~\cite{referit3d}. They serve as the prompt input to the language-grounded perception models for performing 3D visual grounding. Due to increased object density after annotation, identifying unique objects became more challenging. To overcome this, we combined multiple spatial-relationship prompts to exclusively ground objects. See more samples in the appendix.

\begin{figure*}
\vspace{-1ex}
\begin{center}
\includegraphics[width=1.0\linewidth]{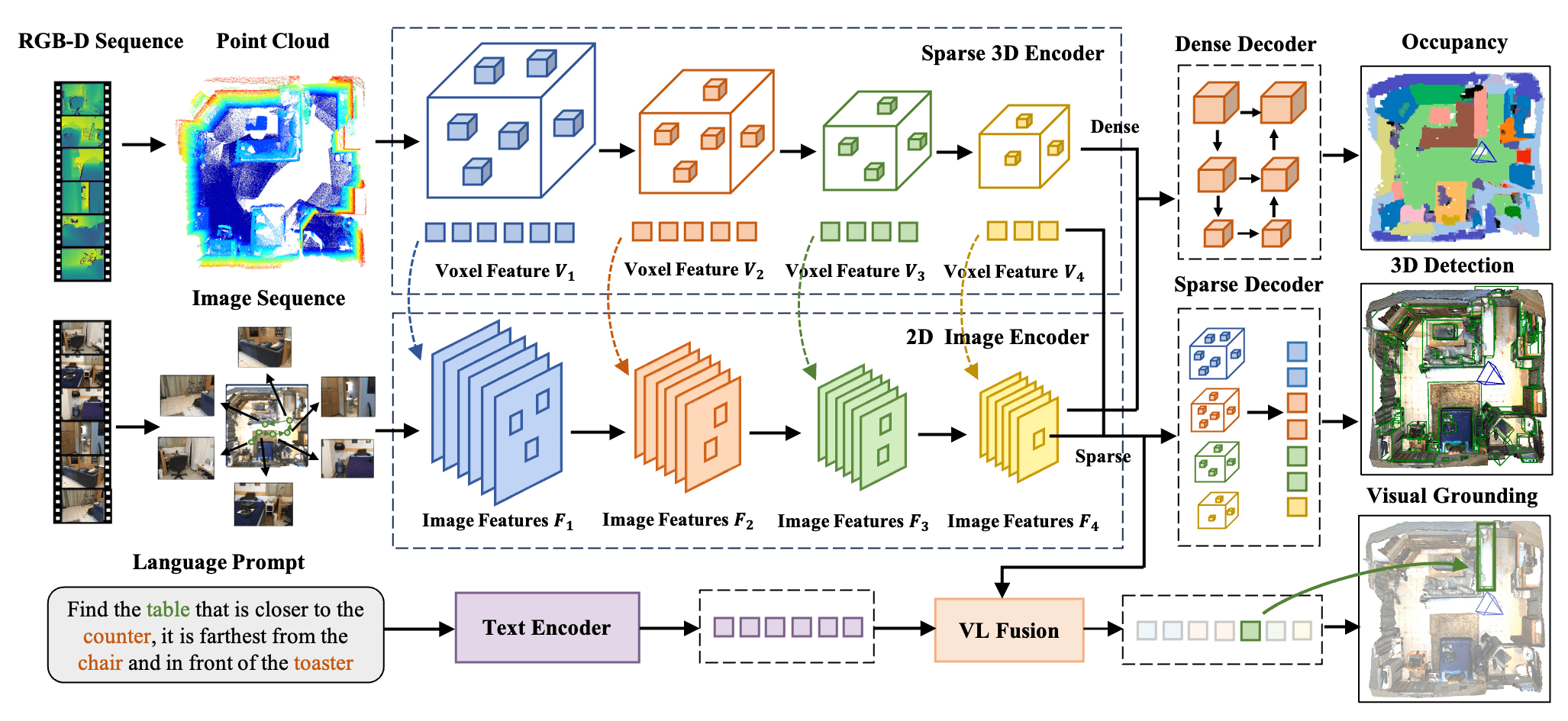}
\end{center}
   \vspace{-4ex}
   \caption{Embodied Perceptron accepts RGB-D sequence with any number of views along with texts as multi-modal input. It uses classical encoders to extract features for each modality and adopts dense and isomorphic sparse fusion with corresponding decoders for different predictions. The 3D features integrated with the text feature can be further used for language-grounded understanding.}
\label{fig: framework}
\vspace{-3ex}
\end{figure*}

\subsection{Statistics}\label{sec:statistics}
\noindent\textbf{Vocabulary Construction.}
During labeling, we ask annotators to write semantic categories in an open-vocabulary manner. This was efficient and suited the complex, large-vocabulary dataset and can provide natural annotations for future open-world research. To sort out these labels, we used Sentence-BERT~\cite{reimers-2019-sentence-bert} to cluster similar categories with text embeddings, match them to WordNet nodes, and finally revise and merge them manually. The vocabulary shares common categories with COCO (50/69 indoor classes) and LVIS (550/1203).

\noindent\textbf{Instance Statistics.} We first show the instances of different categories in Fig.~\ref{fig:instance-stat}.
Our dataset contains over 760 categories, covering common objects in our daily life. More than 288 categories have over 10 instances, and around 400 categories have more than 5 instances. These numbers are 20$\times$ higher than most previous works with 3D box annotations and $3\times$ higher than ScanNet instance segmentation annotations with more than 5 instances. There is also a notable increase in object numbers of small boxes and different categories (Fig.~\ref{fig:3d-box-prompt-stat} and \ref{fig:instance-stat}). We remove four categories, \{wall, ceiling, floor, object\} in our 3D detection benchmark and divide the remaining 284 categories into three splits, \{head, common, tail\} with \{90, 94, 100\} classes.

\noindent\textbf{Occupancy Statistics.}
Semantic occupancy statistics (Fig.~\ref{fig:occ-stat}) reveal the space occupied by different categories, relevant for navigation and motion planning. We chose the first 80 categories for our occupancy prediction benchmark based on distribution and significance in downstream tasks.

\noindent\textbf{Language Prompts Statistics.}
Generated language prompts following SR3D fall into five types of spatial object-to-object relations: Horizontal Proximity, Vertical Proximity, Support, Allocentric, and Between. If a scene has between 2 and 6 instances of a certain category, these categories are considered valid target categories. If a scene has a single instance of a category, it is selected as the anchor category. The training/validation set contains 801711/168322 language prompts, nearly 10 times larger than the original SR3D datasets (Fig.~\ref{fig:3d-box-prompt-stat} and Tab.~\ref{tab:dataset-table}).

\section{Embodied Perceptron}
\label{sec:method}
Given this dataset, we can take multi-modality input, including RGB images, point clouds derived from depth maps as well as language prompts, to extract multi-modal representations and perform different downstream tasks. This section provides a baseline, namely Embodied Perceptron, with a unified framework and customized design for holistic 3D scene understanding from ego-centric views.

\noindent\textbf{Framework Overview.}
The framework includes a multi-modal 3D encoder to extract object \& scene representations and sparse \& dense decoders for various downstream tasks. In addition, we customize the output's parameterization and training objectives to fit the formulation of oriented 3D bounding boxes in the sparse decoder.

\begin{table*}
\scriptsize
\caption{Continuous and multi-view 3D object detection benchmark on EmbodiedScan.}
\vspace{-4.5ex}
\begin{center}
{
     \setlength{\tabcolsep}{3mm}
    \begin{tabular}{c|c|c|c|c|c|cccccc}
    \hline
     \multirow{2}*{Methods} & \multirow{2}*{Input} & \multicolumn{4}{c|}{Large-Vocabulary} & \multicolumn{2}{c}{Head} & \multicolumn{2}{c}{Common} & \multicolumn{2}{c}{Tail}\\
    \cline{3-12}
    ~ & ~ & AP$_{25}$ & AR$_{25}$ & AP$_{50}$ & AR$_{50}$ & AP$_{25}$ & AR$_{25}$ & AP$_{25}$ & AR$_{25}$ & AP$_{25}$ & AR$_{25}$ \\
    \hline
    Camera-Only & RGB & 12.80 & 34.61 & 4.25 & 13.07 & 17.40 & 44.79 & 7.64 & 24.22 & 0.03 & 3.09 \\
    Depth-Only & Depth & 17.16 & 51.40 & 10.52 & 25.75 & 21.39 & 61.14 & 13.27 & 41.58 & 2.74 & 20.91 \\
    Multi-Modality & RGB-D & \textbf{19.07} & \textbf{51.56} & \textbf{11.57} & \textbf{28.15} & 23.54 & 60.23 & 15.80 & 44.99 & 1.24 & 17.74 \\
    \hline\hline
    ImVoxelNet~\cite{imvoxelnet} & RGB & 6.15 & 20.39 & 2.41 & 6.31 & 10.96 & 34.29 & 4.12 & 15.40 & 2.63 & 9.21 \\
    VoteNet~\cite{votenet} & Depth & 3.20 & 6.11 & 0.38 & 1.22 & 6.31 & 12.26 & 1.81 & 3.34 & 1.00 & 1.83 \\
    \hline
    FCAF3D~\cite{fcaf3d} & Depth & 9.07 & 44.23 & 4.11 & 20.22 & 16.54 & 61.38 & 6.73 & 42.77 & 2.67 & 24.83 \\
    +our decoder & Depth & 14.80 & 51.18 & 8.77 & 27.46 & 25.98 & 67.12 & 10.85 & 50.08 & 5.72 & 32.85\\
    +painting & RGB-D & 15.10 & \textbf{51.32} & 8.64 & 26.66 & 26.23 & 67.53 & 11.39 & 50.64 & 5.80 & 32.13 \\
    \hline
    Ours & RGB-D & \textbf{16.85} & 51.07 & \textbf{9.77} & \textbf{28.21} & 28.65 & 67.51 & 12.83 & 50.46 & 7.09 & 31.52 \\
    \hline
    \end{tabular}
}
\vspace{-7ex}
\end{center}
\label{tab:det3d}
\end{table*}

\vspace{-0.5ex}
\subsection{Multi-Modal 3D Encoder}
\vspace{-0.5ex}
As shown in Fig.~\ref{fig: framework}, the multi-modal 3D encoder first has separate encoders for different modalities - ResNet50~\cite{ResNet} and FPN~\cite{FPN} (optional) for 2D images, Minkowski ResNet34~\cite{minkengine} for point clouds, and BERT~\cite{bert} for texts. After extracting these features, we further fuse and process them into sparse or dense features for different downstream tasks. Next, we first present how we aggregate multi-view inputs and then introduce different fusion approaches for dense and sparse feature extraction.

\noindent\textbf{Scalability for Input Views.}
Contrasting with prior works, our framework can accept any number of RGB-D views, making it adaptable and generalizable to varying input orders and quantities. We conveniently aggregate different depth map views by transforming the point clouds into a global coordinate system, downsampling as needed. For multiple images, we query corresponding 2D features using perspective projection from 3D points, averaging them to maintain permutation invariance. This technique allows consistent feature updates during ego-centric exploration.
Theoretically, voxel features could be updated by merging the volume feature at frame $t$ with the incremental feature from RGB-D input at frame $t+1$.
In practice, we accommodate any number of views as batch-wise samples for accelerating training and evaluation. Our model demonstrates notable scalability, where fewer views (e.g., 20) may be used for memory efficiency during training, while more views (\emph{e.g.}, 50) can enhance performance during inference.

\noindent\textbf{Dense Fusion.}
Previous works typically integrate the color and coordinates of points at the input stage, like ``painting"~\cite{pointpainting}, or form multi-modality dense BEV features for concatenated fusion~\cite{bevfusion,liang2022bevfusion}. The latter way suits our occupancy prediction baseline and thus we adopt the straightforward dense fusion on the pre-defined grid, which shares the same resolution with the ground truth. We construct feature volume by projecting grid points onto a 2D feature map post-FPN, then consolidate it with 3D features densified from sparse voxel features. For object detection, we argue that concurrent multi-modality fusion across different feature levels is more effective.

\noindent\textbf{Isomorphic Multi-Level Multi-Modality Fusion.}
Formally, the input aggregated points $P\in \mathcal{R}^{N_p\times 3}$ (first voxelized) and $N_i$ images as $I\in \mathcal{R}^{N_i\times H\times W}$ are processed via a Minkowski ResNet and a shared 2D ResNet respectively. This extracts multi-level sparse voxel features $V_k\in \mathcal{R}^{C_k\times N_{V_k}}$ on $K$ levels and image features $F_s\in \mathcal{R}^{C_s\times H_s\times W_s}$ on $S$ levels. In practice, these two ResNets produce 4 levels of features, for both point clouds and images, denoted as isomorphic multi-modality encoders.


In dense fusion, we filter $F_s$ with an upsampling FPN to derive a feature map $F_{up}$ with $stride=4$ and use it to construct the feature volume for fusing with $V_4$. For the sparse case, we use multi-level features as seeds instead of a single dense feature map to predict 3D objects.
The initial attempt of still query features from $F_{up}$ or raw images $I$ for these seeds is unstable due to inconsistent features for fusion and confusing gradients back-propagation.
Thus, we leverage the isomorphic architecture for level-based projection and feature fusion, \emph{i.e.}, $V_k$ queries the corresponding image features of $F_k$, which empirically shows a better and more stable performance. This method enables multi-level multi-modality feature fusion compared to the ``painting" approach and ensures the consistency of features and gradients across different network levels and modalities.


\noindent\textbf{Vision-Language (VL) Fusion.}
Given the multi-level sparse visual features $F^S_k$ and text features from the text encoder, we use a multi-modal fusion transformer model~\cite{butd-detr,object2scene} for vision-language information interactions. 
Each transformer layer uses a self-attention block to refine sparse visual features and exploit spatial relationships. Then visual and text features interact in cross-modal attention blocks.
This interaction guides updated sparse grounding features $F^G$ to be context-aware for subsequent prediction.

\subsection{Sparse \& Dense Decoder}
Given multi-modal features from typical encoders, we employ separate fusion streams for sparse and dense tasks. This results in four levels of sparse voxel features $F^S_k$ from isomorphic sparse fusion and a single dense feature $F^D$ for decoding and predictions. These are then processed to obtain 3D box and occupancy predictions.

\noindent\textbf{Sparse Decoder for 3D Boxes Prediction.}
Using the multi-level fused features $F^S_k$, we upsample them as in FCAF3D, appending classification, regression, and centerness prediction heads for 3D object detection. In particular, to fit the oriented 3D box output, we add a 6D rotation representation~\cite{6drot} into original regression targets, ultimately decoded as 3D centers $\mathbf{c}$, 3D sizes $\mathbf{l}$, and Euler angles $\mathbf{\Theta}$. Training objectives include the original classification loss, centerness loss, and a disentangled Chamfer Distance (CD) loss for eight corners~\cite{MonoDIS,omni3d}. Specifically, we use one of three groups of decoded predictions, \{3D centers, 3D sizes, and Euler angles\}, while setting the other two with ground truths to compute three corner losses. For example, given 3D sizes and Euler angles ground truth, we can derive the corner loss between the predicted $\mathbf{B}$ and the ground truth box $\hat{\mathbf{B}}$ yielded by 3D center prediction errors:
\vspace{-1ex}
\begin{equation}\label{eqn: 3d_center_corner_loss}
    L_\mathbf{c} = L_{CD}(\mathbf{B}(\mathbf{c}, \hat{\mathbf{l}}, \hat{\mathbf{\Theta}}), \hat{\mathbf{B}})
    \vspace{-1ex}
\end{equation}
Together with the corner loss derived by the overall predicted bounding boxes $L_{pred}$, We balance these losses with preset weights and use them to replace the original box loss:
\vspace{-1.5ex}
\begin{equation}
    L_{loc} = \lambda_\mathbf{c}L_\mathbf{c}+\lambda_\mathbf{l}L_\mathbf{l}+\lambda_\mathbf{\Theta}L_\mathbf{\Theta}+\lambda_{pred}L_{pred}
    \vspace{-0.5ex}
\end{equation}
We set $\lambda_\mathbf{c}=\lambda_\mathbf{l}=\lambda_\mathbf{\Theta}=0.2$ and $\lambda_{pred}=0.4$ to highlight the importance of the overall prediction, which performs well empirically.
The target assignment strategy and post-processing during inference also follow FCAF3D~\cite{fcaf3d}.

\noindent\textbf{Dense Decoder for Occupancy Prediction.}
With the dense feature $F^D$, we use a 3D FPN~\cite{imvoxelnet} to aggregate multi-level features and produce multi-scale occupancy predictions. Since the task requires more powerful low-level features for fine detail understanding, predictions at each scale are thus supervised with decayed half weights from high to low resolution~\cite{monoscene}. We use cross-entropy loss and scene-class affinity loss~\cite{surroundocc} for training. During inference, we only use the high-resolution output for prediction.

\noindent\textbf{Sparse Decoder for 3D Visual Grounding}.
Grounding features $F^G$ updated after each transformer layer are fed into the prediction heads sharing the same architecture as those used for 3D detection.
All prediction head outputs in each layer are supervised during training for stability and improved performance. An additional contrastive loss aligns the visual feature with target text prompts, ensuring the features of a target text token are closer to corresponding visual features and further from other visual or text tokens.


\begin{table*}[t]
\scriptsize
\setlength{\tabcolsep}{3.5pt}
\caption{Continuous and multi-view occupancy prediction benchmark on EmbodiedScan. ``refri." means ``refrigerator".}
\vspace{-2ex}
\centering
{%
    \begin{tabular}{c|c|c|cccccccccccccccc}
    \hline
    Methods  & Input & mIOU  & empty & floor &  wall & chair & cabinet & door & table & couch & shelf & window & bed & curtain & refri. & plant & stairs & toilet \\
    \hline
    Camera-Only & RGB & 10.43 & 39.09 & 34.10 & 30.24 & 26.46 & 9.49 & 25.53 & 41.60 & 35.19 & 16.22 & 20.45 & 24.46 & 19.80 & 26.01 & 17.26 & 1.83 & 29.78 \\
    Depth-Only & Depth & 14.44 & 73.91 & 66.22 & 56.13 & 49.96 & 15.70 & 24.37 & 56.84 & 55.35 & 30.55 & 26.66 & 42.81 & 30.81 & 33.01 & 21.71 & 6.21 & 45.35 \\
    Multi-Modality & RGB-D & \textbf{20.79} & 73.50 & 63.64 & 62.30 & 54.60 & 19.96 & 48.99 & 61.10 & 69.76 & 39.86 & 34.62 & 54.83 & 54.45 & 48.90 & 41.22 & 7.97 & 63.52 \\
    \hline\hline
    OccNet~\cite{occnet} & RGB & 8.07 & 37.15 & 46.90 & 25.63 & 20.94 & 13.17 & 18.40 & 26.81 & 22.86 & 13.59 & 13.49 & 26.75 & 22.92 & 17.15 & 17.07 & 4.77 & 33.60 \\
    SurroundOcc~\cite{surroundocc} & RGB & 9.10 & 38.54 & 46.17 & 23.55 & 23.04 & 13.60 & 19.15 & 27.79 & 22.88 & 13.11 & 13.72 & 24.32 & 18.89 & 13.58 & 14.77 & 7.83 & 34.71 \\
    \hline
    Camera-Only & RGB & 10.48 & 40.45 & 41.25 & 27.19 & 26.16 & 15.50 & 20.30 & 30.82 & 26.70 & 15.01 & 14.33 & 29.17 & 23.30 & 16.99 & 15.98 & 6.17 & 42.57 \\
    Depth-Only & Depth & 15.56 & 69.92 & 60.52 & 51.74 & 49.44 & 23.08 & 24.33 & 45.77 & 43.52 & 29.74 & 23.02 & 39.04 & 41.22 & 17.42 & 19.58 & 25.79 & 60.45 \\
    Multi-Modality & RGB-D & \textbf{19.97} & 71.21 & 64.92 & 55.00 & 52.04 & 27.35 & 33.97 & 47.93 & 46.26 & 31.87 & 27.98 & 46.58 & 46.56 & 24.05 & 39.01 & 24.40 & 67.79 \\
    \hline
    \end{tabular}%
}
\vspace{-2ex}
\label{tab:occ}
\end{table*}

\begin{table*}
\scriptsize
\caption{Monocular 3D object detection benchmark on EmbodiedScan.}
\vspace{-5ex}
\begin{center}
{
     \setlength{\tabcolsep}{2.5mm}
    \begin{tabular}{c|c|c|c|c|c|cccccccc}
    \hline
     \multirow{2}*{Methods} & \multirow{2}*{Input} & \multicolumn{4}{c|}{20 Common Classes} & \multirow{2}*{chair} & \multirow{2}*{cabinet} & \multirow{2}*{table} & \multirow{2}*{bin} & \multirow{2}*{couch} & \multirow{2}*{bed} & \multirow{2}*{bathtub} & \multirow{2}*{toilet}\\
    \cline{3-6}
    ~ & ~ & AP$_{25}$ & AR$_{25}$ & AP$_{50}$ & AR$_{50}$ & ~ & ~ & ~ & ~ & ~ & ~ & ~ & ~ \\
    \hline
    FCOS3D~\cite{FCOS3D} & RGB & 8.93 & 27.96 & 0.91 & 5.00 & 27.15 & 1.14 & 6.21 & 10.23 & 9.47 & 18.38 & 6.31 & 40.51  \\
    ImVoxelNet~\cite{imvoxelnet} & RGB & 18.95 & 52.74 & 1.81 & 7.10 & 46.70 & 4.63 & 18.10 & 17.82 & 20.39 & 41.51 & 10.14 & 65.70 \\
    VoteNet~\cite{votenet} & Depth & 14.30 & 31.44 & 1.68 & 5.14 & 54.00 & 2.41 & 19.53 & 14.72 & 21.80 & 45.58 & 13.49 & 68.16 \\
    ImVoteNet~\cite{imvotenet} & RGB-D & 19.63 & 34.32 & 3.88 & 8.82 & 56.72 & 2.88 & 29.00 & 21.96 & 27.77 & 56.94 & 37.56 & 74.08 \\
    \hline
    FCAF3D~\cite{fcaf3d} & Depth & 25.70 & 78.53 & 5.73 & 20.26 & 65.91 & 6.47 & 26.64 & 34.93 & 22.50 & 53.68 & 26.38 & 71.90 \\
    +our decoder & Depth & 28.16 & 84.50 & 4.92 & 20.69 & 63.85 & 6.62 & 32.34 & 38.96 & 31.61 & 60.33 & 38.17 & 75.57\\
    +painting & RGB-D & 30.19 & 83.93 & 5.74 & 21.90 & 66.39 & 7.41 & 33.66 & 42.86 & 32.24 & 60.04 & 41.31 & 77.59 \\
    \hline
    Ours & RGB-D & \textbf{34.28} & \textbf{85.03} & \textbf{12.61} & \textbf{32.25} & 69.47 & 10.01 & 37.29 & 45.17 & 31.67 & 63.27 & 50.63 & 80.39 \\
    \hline
    \end{tabular}
}
\vspace{-6ex}
\end{center}
\label{tab:mono3d}
\end{table*}

\section{Benchmark}\label{sec:benchmark}


Our benchmark has three categories based on data samples: scene-based, view-based, and prompt-based. Scene-based benchmarks mean the samples are based on different scenes, covering continuous and multi-view perception. View-based benchmarks use ego-centric views for tasks like monocular 3D detection. Lastly, samples of 3D visual grounding are based on constructed language prompts. Detailed splits will be discussed in each benchmark.

For metrics, we use the 3D IoU-based average precision (AP) with thresholds of 0.25 and 0.5 for 3D detection and visual grounding. We also provide average recall (AR) for reference. For occupancy prediction, we employ the mean Intersection of Union (mIoU) as a performance measure. Due to the space limitation, please refer to the appendix for implementation details of different baselines, and more quantitative and qualitative results including an "in-the-wild" evaluation demo.



\subsection{Fundamental 3D Perception Benchmarks}
\noindent\textbf{Continuous 3D Perception.}
As opposed to driving scenarios, indoor scene understanding is typically in an enclosed space, making it important to fully leverage multi-view cues formed by RGB-D sequence and continuously maintain an overall scene-level representation.
Thus, we design this new benchmark involving sequential views for perceiving covered 3D regions. Models are trained and evaluated scene-wise with 3930/703/552 scans allocated for training/validation/testing.
To accelerate the training and evaluation, we construct $N$ data samples with $1\sim N$ views from $N$ sampled views per scan. Here, $N=10$ during training with random view sampling, while in evaluation, $N=50$ with fixed views.
Corresponding instances and occupancy truths are obtained by combining pre-computed visible instance IDs and occupancy masks of selected views. If a category lacks instances, it is removed when calculating mAP and mIoU. Given this new setup, we primarily offer three baselines with different input modalities (Tab.~\ref{tab:det3d} and \ref{tab:occ}).

\noindent\emph{Continuous 3D Object Detection.}
As anticipated, both RGB and depth features significantly impact this task, leading to superior results of our RGB-D approach (Tab.~\ref{tab:det3d}). The performance of the depth-only model closely mirrors the multi-modality approach, indicating depth's dominance in 3D perception. Our method of constructing multi-modal features based on sparse voxel features also aligns with this intuition. Low performance on tail categories suggests dataset size influences performance, warranting future enhancement.

\noindent\emph{Continuous Semantic Occupancy Prediction.}
This benchmark offers comprehensive results including mIoU and IoU for common classes. Unlike the detection benchmark, there is a notable gap between the depth-only and RGB-D baseline. This might be due to the former's limited semantic understanding capability, especially evident in categories like door and curtain, which are similar to walls in shape.
On such a task that requires more fine-grained understanding, the depth sensor's weakness is enlarged.
Meanwhile, depth plays a crucial role in predicting empty space, floor, and wall, while RGB information substantially improves prediction for most categories.

\begin{table}
\scriptsize
\caption{Multi-view 3D visual grounding benchmark. ``Indep/Dep" refer to ``View-Independent/Dependent". Easy/Hard and Indep/Dep have a ratio of 80\%/20\% and 78\%/22\%.}
\vspace{-5ex}
\begin{center}
{
    \begin{tabular}{c|c|c|cc|cc}
    \hline
     \multirow{2}*{Methods} & \multirow{2}*{Input} & \multicolumn{1}{c|}{Overall} &
     \multicolumn{1}{c}{Easy} & \multicolumn{1}{c|}{Hard} & \multicolumn{1}{c}{Indep} & \multicolumn{1}{c}{Dep}
     \\
    \cline{3-7}
    ~ & ~ & AP$_{25}$ & AP$_{25}$ & AP$_{25}$ & AP$_{25}$ & AP$_{25}$\\
    \hline
    ScanRefer~\cite{scanrefer} & RGB-D & 12.85 & 13.78 & 9.12 & 13.44 & 10.77 \\
    BUTD-DETR~\cite{butd-detr} & RGB-D & 22.14 & 23.12 & 18.23 & 22.47 & 20.98 \\
    L3Det~\cite{object2scene} & RGB-D & 23.07 & 24.01 & 18.34 & 23.59 & 21.22 \\
    \hline
    Ours & RGB-D & \textbf{25.72} & \textbf{27.11} & \textbf{20.12} & \textbf{26.37} & \textbf{23.42} \\
    \hline
    \end{tabular}
}
\vspace{-8ex}
\end{center}
\label{tab:vg_3d}
\end{table}

\noindent\textbf{Multi-View 3D Perception.}
Unlike continuous settings, multi-view 3D perception does not predefine the order of views but provides all views to the model for scene-level results. This setting was studied previously~\cite{imvoxelnet}, so we first reproduce common methods on our benchmark.

\noindent\emph{Multi-View 3D Object Detection.}
We implement baselines including ImVoxelNet~\cite{imvoxelnet} with RGB-only input and VoteNet~\cite{votenet} and FCAF3D~\cite{fcaf3d} with depth-only input (Tab.~\ref{tab:det3d}). Additional dimensions are added to predict Euler angles with a simple L1 loss on their cosine values, but it yields underwhelming results. Substituting this with our decoder design markedly improves performance. Further using point cloud input painted for FCAF3D with RGB-D input slightly underperforms our baseline. Nevertheless, all models have substantial potential for improvement, demonstrating the challenges of this new dataset and setup.

\noindent\emph{Multi-View Semantic Occupancy Prediction.}
We implement two popular baselines from autonomous driving benchmarks, OccNet~\cite{occnet} and SurroundOcc~\cite{surroundocc} (similar to TPVFormer~\cite{tpvformer}). Their performance slightly lags behind our camera-only baseline. Variants of our baselines exhibit a performance trend akin to embodied benchmarks.

\noindent\textbf{Monocular 3D Perception.}
Finally, the basic ego-centric setting is monocular 3D perception, specifically 3D detection, where each data sample comprises a single RGB-D frame and corresponding visible 3D boxes. Scan splits are used to extract frames as data samples, resulting in 689k/115k/86k images for the training/validation/testing.

\noindent\emph{Monocular 3D Object Detection.}
This is more challenging than multi-view due to the absence of stereo geometric cues and truncated object views in indoor scenes, so the performance is significantly reduced in large-vocabulary settings. Hence, we first create a benchmark for 20 common categories (Tab.~\ref{tab:mono3d}), observing a larger AP-AR gap for top methods because of difficulties predicting accurate 3D boxes from partial views. Similarly, our method outperforms others, providing a solid baseline for future studies.

\begin{table}
\scriptsize
\caption{Ablation with conventional settings.}
\vspace{-5ex}
\begin{center}
{
    \begin{tabular}{c|c|c|c|c|c}
    \hline
    Oriented & Multi-View & AP$_{25}$ & AR$_{25}$ & AP$_{50}$ & AR$_{50}$ \\
    \hline
    \ding{55} & \ding{55} & 70.17 & 90.46 & 54.58 & 75.66 \\
    \ding{51} & \ding{55} & 61.87 & 90.31 & 47.30 & 73.93 \\
    \ding{51} & \ding{51} & 59.95 & 87.92 & 43.33 & 69.95 \\
    \hline
    \end{tabular}
}
\vspace{-5ex}
\end{center}
\label{tab:connect_with_conventional}
\end{table}

\begin{table}
\scriptsize
\caption{Real vs. rendered images on ScanNet.}
\vspace{-5ex}
\begin{center}
{
    \begin{tabular}{c|c|c|ccc}
    \hline
     Train & Val & Overall & Head & Common & Tail \\
    \hline
    Render & Render & 22.11 & 33.01 & 16.44 & 6.74 \\
    Render & Real & 18.72 & 27.02 & 14.85 & 6.25 \\
    Real & Real & 21.98 & 32.91 & 17.18 & 5.05 \\
    \hline
    \end{tabular}
}
\vspace{-7ex}
\end{center}
\label{tab:real_vs_render}
\end{table}

\subsection{Language-Grounded Benchmark}
\noindent\textbf{Multi-View 3D Visual Grounding.}
Our benchmark introduces language into the perception loop to foster interactive 3D scene representation learning. With comprehensive instance annotations, our benchmark presents more complex prompts and grounding cases than previous works. As an initial step, this setup takes multi-view RGB-D images as input without considering differing prompt timestamps. The goal is to ground the object described by the language prompt in the scene using information from different ego-centric views. Ground-truth detection boxes are not provided as candidates for grounding during evaluation, which can better validate end-to-end 3D visual grounding ability than the original SR3D~\cite{referit3d}. Data sample splits align with previous benchmarks' 3D scan splits.

We reimplement classic methods like ScanRefer~\cite{scanrefer}, BUTD-DETR~\cite{butd-detr}, and L3Det~\cite{object2scene} (Tab.~\ref{tab:vg_3d}). Our baseline outperforms all due to the strong multi-modal encoder. However, the performance remains much lower than previous works, partly due to changes in input format and annotations, which we will analyze next. Further challenges arise from handling more categories and small objects, making the grounding task more complex in parsing input prompts and predictions. Addressing these new challenges in this classical task would be promising for future research.


\subsection{Analysis}
\label{sec:analysis}
Finally, we make further analysis to connect EmbodiedScan to current progress in computer vision.  

\noindent\textbf{Axis-aligned vs. Oriented Boxes.}
We start with the 18-class detection performance of FCAF3D on ScanNet (Tab.~\ref{tab:connect_with_conventional}). First, we change the annotations to oriented 3D boxes and adapt with our decoder. We find a significant drop in performance, indicating that the orientation estimation makes this task more challenging. We need to explore a better method to represent and predict the object pose.

\noindent\textbf{Reconstructed Point Cloud vs. Multi-View RGB-D.}
Subsequently, replacing the reconstructed point clouds with the aggregated ones from multi-view depth maps has minor effects on $AP_{25}$ but heavily impacts $AP_{50}$ (Tab.~\ref{tab:connect_with_conventional}), implying that the accuracy of reconstructed point clouds is superior to raw depth maps. Therefore, integrating reconstruction techniques in perception loops shows potential.

\begin{table}
\scriptsize
\caption{Benefits from training with EmbodiedScan.}
\vspace{-4.5ex}
\begin{center}
{
    \begin{tabular}{c|c|c|ccc}
    \hline
     Train & Val & Overall & Head & Common & Tail \\
    \hline
    ScanNet & ScanNet & 20.28 & 29.81 & 15.57 & 6.40 \\
    Ours & ScanNet & \textbf{23.02} & 33.82 & 18.09 & 6.57 \\
    \hline
    ScanNet & Ours & 10.92 & 21.10 & 8.06 & 1.78 \\
    Ours & Ours & \textbf{16.85} & 28.65 & 12.83 & 7.09 \\
    \hline
    \end{tabular}
}
\vspace{-8ex}
\end{center}
\label{tab:embodiedscan_training}
\end{table}

Next, we study the gap between the real and rendered images, and the benefits from training with EmbodiedScan. We do the comparison with our multi-modality baseline on the large-vocabulary multi-view 3D detection benchmark.

\noindent\textbf{Real Capture vs. Rendering.}
As shown in Tab.~\ref{tab:real_vs_render}, apart from the significant visual difference between real and rendered images (Fig.~\ref{fig:teaser}), the model's performance also has a remarkable decrease when transferring models trained with rendered images to the real world, particularly when the annotations are sufficient (5.99\% AP drop on head categories and 5.89\% AP lower than models trained with real images.)

\noindent\textbf{Benefits from EmbodiedScan.} Finally, we also quantitatively evaluate the benefits of training models with our large-scale EmbodiedScan (Tab.~\ref{tab:embodiedscan_training}). As expected, when training our models with EmbodiedScan, we observed a significant improvement in both ScanNet (2.74\% AP) and the overall validation split (5.93\% AP), particularly 4.01\% AP and 7.55\% AP increase on head categories. 

\vspace{-0.5ex}
\section{Conclusion}
\label{sec:conclusion}
This paper introduces EmbodiedScan, a multi-modal perception suite aiming for language-grounded holistic 3D scene understanding from ego-centric views. We construct a large-scale dataset with diverse sensor data and multi-modal annotations, including 3D oriented boxes, semantic occupancy and language descriptions. Based on this dataset, we propose a baseline framework capable of handling any number of views input, using a unified multi-modal encoder and task-specific decoders. We establish benchmarks for basic and language-grounded 3D perception. Experiment results highlight our work's value and reveal new challenges in this setup. We believe EmbodiedScan can bring opportunities in embodied 3D perception and may also have a broader impact in related fields with the massive data and rich annotations.

\appendix
\section{Implementation Details}
In the main paper, we focus on the overall design of our baseline framework and the benchmark results.
Here, we further provide the implementation details of Embodied Perceptron and other baselines on our benchmark.
\subsection{Embodied Perceptron}
\noindent\textbf{Input.}
Except for the monocular task, given the memory usage of different tasks, we set different numbers of input images during training and inference. Specifically, we set the number of input images to 20 and 50 for training and inference of multi-view 3D detection and visual grounding experiments while reducing the number to 10 for training continuous 3D detection models.
For occupancy experiments, we set the number to 10 and 20 for training and inference.
In addition, due to different resolutions of images from different source datasets, we resize them to $480\times 480$ for unification and conduct corresponding transformations when computing the projection from points to images.

For depth maps, after converting them to point clouds, we first sample the points to limit their maximum number to 100k~\cite{fcaf3d}. Then we voxelize them and feed them into the sparse convolutional networks. We set the voxel size to 0.01 meters for 3D detection following the convention of previous works~\cite{fcaf3d}. In contrast, since we only use the last-level voxel feature (64$\times$ downsampled) to construct the feature volume, the voxel size is set to 0.16/64=0.0025 meters to ultimately predict the $40\times 40\times 16$ occupancy (the output voxel size is 0.16 meters).

\noindent\textbf{Multi-Modal 3D Encoder.}
As mentioned in the main paper, we use the classical encoders for different modalities at the beginning, \emph{i.e.}, a shared ResNet50~\cite{ResNet} for multi-view images, MinkResNet34~\cite{minkengine} for point clouds derived from depth maps, and RoBERTa-Base~\cite{roberta} for texts. Here, we reduce the base channels in ResNet to 16 to make it consistent with MinkResNet34, resulting in \{128, 256, 512, 1024\} multi-level feature channels after sparse fusion. In contrast, for dense fusion, we keep the original setting of ResNet, use FPN to enhance the 2D features (256 channels) to derive the 3D feature volume, and finally concatenate it with the densified last-level voxel feature $V_4$ (512 channels), resulting in 768-channel dense feature for subsequent occupancy prediction. For different outputs, the current encoder design has shared separated encoders but minor differences during fusion. How to further unify them and how it could benefit multi-task training and pre-training would be intriguing problems to be explored in the future.

\noindent\textbf{Spare \& Dense Decoder.} We basically follow FCAF3D~\cite{fcaf3d} in 3D detection head designs but adapt it to be compatible with oriented 3D boxes. It generates predictions based on sparse voxel seeds and assigns targets to them according to several rules during training, such as whether the voxel center is inside a box and assigning it to the best feature level similar to FCOS~\cite{FCOS,FCOS3D}. Please see more details in its original paper.
Here, all the computations regarding the distance between centers, points, and six faces and box formulations mentioned in the main paper are modified to fit the 3-DoF rotation version.

For the dense decoder, we first use a 3D FPN~\cite{imvoxelnet} to filter the 3D dense feature and compress the feature channel to 128. The output multi-level features in three resolutions, from $40\times40\times16$ to $10\times10\times8$, are fed into three occupancy prediction heads, which share the same architecture, a 3D convolutional layer with kernel size and stride set to 1, to produce the multi-scale results. The training objective and loss are similar to SurroundOcc~\cite{surroundocc} for supervising the multi-scale output. During inference, we only take the high-resolution output as the final prediction.

For the visual grounding decoder, we adopt several transformer layers to fuse the 3D sparse feature and text feature. Similar to GroupFree3D~\cite{groupfree3d}, we refine the position encoding of an object candidate stage by stage. Specifically, we predict the 3D box locations at each decoder layer, and the predicted box location will be used to produce the updated position encoding of the same query. The queries are updated iteratively through $N_{D}$ = 6 decoder layers. Besides, to achieve the contrastive loss mentioned in our original paper, one visual projection layer and one text projection layer are needed to project visual and text features to the same feature space with channel 64 for alignment. The projection layer consists of three linear layers. The contrastive loss aims to learn the similarity of visual-text multimodal features, consisting of two losses:  $\mathcal{L}_{con}^v$ ensures the features of an object query are closer to positive text token features and farther from other text tokens, and $\mathcal{L}_{con}^t$ ensures
the features of a target text token are closer to corresponding visual features and farther from other visual tokens.

\begin{small}
\begin{equation}
    \mathcal{L}_{con}^v\! = \! \sum_{i=1}^{k} -\log \!\left(\!\frac{\exp \left(\boldsymbol{o}_{i}^{\top} \boldsymbol{t}_{i} / \tau\right)}{\sum_{j=1}^{l} \exp \left(\boldsymbol{o}_{i}^{\top} \boldsymbol{t}_{j} / \tau\right)}\!\right),
    \label{eq:loss_con_v}
\end{equation}
\end{small}

\begin{small}
\begin{equation}
    \mathcal{L}_{con}^t = \sum_{i=1}^{l} -\log \left(\frac{\exp \left (\boldsymbol{t}_{i}^{\top} \boldsymbol{o}_{i} / \tau\right)}{\sum_{j=1}^{k} \exp \left (\boldsymbol{t}_{i}^{\top} \boldsymbol{o}_{j} / \tau \right)}\right),
    \label{eq:loss_con_t}
\end{equation}
\end{small}

\begin{small}
\begin{equation}
    \mathcal{L}_{con} = \mathcal{L}_{con}^v + \mathcal{L}_{con}^t,
    \label{eq:loss_con}
\end{equation}
\end{small}

where $\boldsymbol{o}$ and $\boldsymbol{t}$ are the object and text features after projection layers, and $\boldsymbol{o}^{\top} \boldsymbol{t} / \tau$ is their similarity.
$k$ and $l$ are the number of objects and words.
$\boldsymbol{t}_i$ is the positive word feature of the i-th candidate object.


\noindent\textbf{Training Parameters.}
For all the experiments, we only use the pre-trained ResNet50 and RoBERTa provided by PyTorch while training other modules randomly initialized from scratch following end-to-end manners. The network is trained using AdamW~\cite{adamw} optimizer, with $\beta_1=0.9$, $\beta_2=0.999$. For continuous/multi-view/monocular 3D detection, we use 8 GPUs with 1/4/8 training samples on each to train the model for 96/120/24 epochs, setting the learning rate to 0.0002/0.001/0.0002 and weight decay to 0.0001. For all the occupancy experiments, we use 8 GPUs with 1 training sample on each to train the model for 24 epochs, setting the learning rate to 0.0001 and weight decay to 0.01.

\noindent\textbf{Data Augmentation.} Since the transformation of 3D boxes is easier than occupancy, we only conduct input data augmentations for 3D detection experiments. For continuous and multi-view 3D detection, we randomly flip and apply global transformations to the aggregated points, including random rotation with angles in $[-0.0873, 0.0873]$, random scaling with a ratio in $[0.9, 1.1]$ and random translation following a normal distribution with standard deviation 0.1, but do not apply any augmentation to images.
The rotate and flip augmentations are removed for the view-dependent 3D visual grounding experiments.
For monocular experiments, we use the same augmentation settings while also flipping 2D images for better performance.

\subsection{3D Detection Baselines}
By default, our following re-implemented baselines use the same input setup (such as the number of input views) and backbone with Embodied Perceptron for consistent performance comparison, \emph{e.g.}, ResNet50 for ImVoxelNet and FCOS3D, MinkResNet34 for FCAF3D, \emph{etc.} All the baseline implementation starts with a basic adaptation for oriented 3D box prediction, a simple L1 loss on the Euler angles' cosine values, and can smoothly change the decoder formulation to ours for performance improvement, as mentioned in the main paper. We basically provide several key hyperparameters as follows and all the models are trained to fully converge.
The re-implementation of baselines is based on their official release on top of MMDetection3D~\cite{mmdet3d2020} and more details can be referred to our code release.

\noindent\textbf{ImVoxelNet.}
We adapt its officially released code to fit our dataset and experiments. For multi-view 3D detection, we set the grid range to $[-3.2m\sim 3.2m, -3.2m\sim 3.2m, -0.78m\sim 1.78m]$ along the X-Y (horizontal) plane and Z (vertical) axis and adopt a random origin shift augmentation following a normal distribution with standard deviation $[0.7, 0.7, 0]$ along these axes as in the original paper.
We use 8 GPUs with 1 training sample on each to train the model for 36 epochs, setting the learning rate to 0.0001 and weight decay to 0.0001.
For monocular 3D detection, we change the range to $[-3.2m\sim 3.2m, -1.0m\sim 1.56m, 0.8m\sim 7.2m]$ along the XYZ axis where Y and Z correspond to the height and depth axis. We use 8 GPUs with 4 samples on each to train the model for 12 epochs, setting the learning rate to 0.0002 and the weight decay to 0.0001.
All the backbones would use a $0.1\times$ smaller learning rate as in the original implementation.

\noindent\textbf{VoteNet.} We follow the official version for 3D detection on ScanNet and only change the orientation estimation to the trivial version mentioned above. We derive the mean 3D size of boxes according to our annotations for the partial bin-based box coder setup. We use 8 GPUs with 4/8 samples on each to train the model for 180/12 epochs for multi-view/monocular 3D detection and adopt data augmentations for point clouds including random flip, rotation, scaling, and translations, similar to our baseline. It was observed that the performance decreased a lot when changing the original multi-bin orientation estimation to the trivial one. It is also challenging for the original classification and localization design for our large-vocabulary setting. Therefore, how to design a more effective head for such point-based methods is an important problem to be explored afterward.

\noindent\textbf{ImVoteNet.} Since adapting ImVoteNet for multi-view cases is non-trivial, we only implement it for monocular experiments for comparison. Following the official implementation, we first train a Faster R-CNN with the amodal 2D boxes derived from projected 3D boxes in the first stage and then tune the overall framework in the second stage. Related hyperparameters are set similar to VoteNet. We use 8 GPUs with 2/16 training samples on each to train the model for 8/24 epochs at the first/second stage, setting the learning rate to 0.02/0.001 and the weight decay to 0.0001/0.01 with the SGD/AdamW optimizer.

\noindent\textbf{FCAF3D.} FCAF3D is similar to our depth-only baseline, with differences in network designs such as multi-modality fusion and decoders. Thus, the hyperparameters are also set similarly to our baselines, including the input, optimizers, training epochs, and data augmentations.

\noindent\textbf{FCOS3D.} FCOS3D is a conventional baseline for monocular 3D detection with simple architecture designs. We follow the official implementation but change the backbone to ResNet50 for consistency. We use 8 GPUs with 8 training samples on each to train the model for 24 epochs, setting the learning rate to 0.024 and the weight decay to 0.0001 with the SGD optimizer.

\subsection{Occupancy Prediction Baselines}
We follow the original implementation of OccNet~\cite{occnet} and SurroundOcc~\cite{surroundocc} but change the BEV query according to our settings to finally derive the $40\times 40\times 16$ occupancy in the range $[-3.2m\sim 3.2m, -3.2m\sim 3.2m, -0.78m\sim 1.78m]$. Besides, we take the RGB-D sequence input in our setting as the multi-camera input in its original paper (for autonomous driving). The learning rate is set to 0.0002/0.0001 and the weight decay is set to 0.01/0.01 for the AdamW optimizer of OccNet and SurroundOcc. We use 8 GPUs with 1 sample on each to train the model for 48 epochs. Following the official code, we also use a $0.1\times$ smaller learning rate for their backbone weights update.

\subsection{Visual Grounding Baselines}
\noindent\textbf{ScanRefer.}
We implement ReferNet~\cite{scanrefer} based on our adapted VoteNet and do not change other designs. We use 4 GPUs with 14 training samples on each to train the model for 48 epochs. The learning rate is set to be 1e-3 and the weight decay is set to 1e-5.

\noindent\textbf{BUTD-DETR.}
We reimplement the BUTD-DETR in our codebase and also change the orientation estimation to achieve oriented 3D box prediction. Considering the large vocabulary setting in our benchmark, we do not predict the residual size of the 3D boxes based on their mean sizes calculated according to the annotations. Instead, following the original setting of BUTD-DETR, the essence of the visual grounding task is not to predict the 3D box and its category, but to predict the alignment score between one 3D box and the input prompt. We directly predict the actual size of each 3D box.
Besides, we keep the input of the box stream unchanged as in its official implementation, \emph{i.e.}, a pre-trained GroupFree3D detector is used to obtain 3D object box proposals, which are sent into the box stream.
We use 4 GPUs with 24 training samples on each to train the model for 80 epochs. The learning rate of the backbone is set to be 1e-5, the text encoder is frozen and the learning rate of the remaining parts is set to be 1e-4.

\noindent\textbf{L3Det.} L3Det is a cleaner architecture that is modified based on BUTD-DETR, where the text and visual feature fusion is conducted in the decoder. Similar to BUTD-DETR, we change the orientation estimation to adapt to the oriented 3D box prediction, and other components remain unchanged. We modify BUTD-DETR and reimplement L3Det~\cite{object2scene} as a cleaner architecture on top. The settings for training and optimization are the same as those of BUTD-DETR. 

\section{Dataset Details}
\subsection{Data Processing}
\noindent\textbf{Difference Among Source Datasets.}
In the main paper, we mentioned that although the source datasets all have RGB-D data, their data distributions have significant differences.
Specifically, ScanNet provides the raw RGB-D sequence with the most frames (highest sampling frequency) and the image resolution $1296\times 968$. 3RScan uses a portrait screen with image resolution $540\times 960$ (but provides the image with $90^\circ$ rotation, resulting in $960\times 540$) and has fewer frames. Matterport3D directly provides general multi-view images with image resolution $1280\times 1024$ instead of video sequences to serve as the image modality of 90 building-scale scenes. In the main paper, we have mentioned that we unify the input as a general multi-view case and sample ScanNet frames to make them consistent with the other two datasets. In addition, we rescale the images to $480\times 480$ to extract 2D features to unify the resolutions of inputs and also force the 2D backbone to learn features robust to the scale and rotations.

\subsection{Annotation}
\noindent\textbf{Definition of Oriented 3D Boxes.} As mentioned in the main paper, we follow the typical definition of oriented 3D boxes, including 3D center, 3D size, and three Euler angles. This definition is naturally transferred from previous research in 3D detection, from 7-DoF boxes in autonomous driving to this 9-DoF version for any 3-DoF rotation. However, it still has some ambiguity in the definition of 3D sizes and orientations. Because the definition of length, width, and height is ambiguous, we define the 3D size as the length along the XYZ axis, $\Delta x$, $\Delta y$, $\Delta z$, to constitute the 3D size. A potentially tricky problem is that we may have multiple solutions with different combinations of this 3D size and Euler angles for a specific oriented box. This is because we do not pre-define the 3D size for each object to just estimate the orientation, which is more similar to the settings of 6D pose estimation. As a result, our 9-DoF definition is more suitable for the detection setting considering that it should be more general for different objects from the large-vocabulary categories, but at the same time essentially can be reduced to a definition of boxes with eight vertices and a single normal/unidirectional orientation. It may lose other dimensions of orientation (compared to 6D pose) information, but in practice, such a unidirectional orientation is enough for most objects, considering many of them are symmetric. The discussion about such object representations and the corresponding evaluation metric design can be important in future works.

\noindent\textbf{Language Prompt Generation.}
When producing language prompts, each prompt is designed to uniquely identify a target object within a 3D scan by establishing a distinct relationship between the target and an adjacent object, referred to as the ``anchor". Following SR3D~\cite{referit3d}, we use the following compositional template to construct the language prompt:
\[
\langle \text{target class} \rangle\ \langle \text{spatial relation} \rangle\ \langle \text{anchor class(es)} \rangle
\]
We present the five types of spatial-relation language prompts to make this appendix self-contained: Horizontal Proximity, Vertical Proximity, Support, Allocentric, and Between~\cite{referit3d}:

\begin{itemize}
    \item \textbf{Horizontal Proximity:} The type of language prompt shows the distance in the horizontal direction between the target and anchor objects, which indicates how close or far the target is from the anchor in the scene.
    \item \textbf{Vertical Proximity:} This prompt indicates the vertical relationship between the target and anchor.
    \item \textbf{Between:} This prompt indicates there exists a target between the two anchors. Furthermore, we can obtain a more precise description, such as the target being in the middle of two anchors.
    \item \textbf{Allocentric:} Based on our new EmbodiedScan annotation, each object will contain precise \emph{orientation} information. Based on the position vector between the target and anchor, as well as the orientation vector of the anchor object itself, we can easily determine whether the target is in front/back/left/right of the anchor.
    \item \textbf{Support:} This prompt indicates that the target is either supported by or supporting the anchor.
\end{itemize}

We generate each type of language prompt scene by scene. Before prompt generation, we need to separately filter out classes that are suitable as targets and anchors. In addition, for each scene, we need to further determine the valid class as the following:

\begin{itemize}
    \item A class is a valid class for a target if: 1) There must be multiple objects of this class in the current scene. 2) The number of objects of this class in the current scene cannot exceed 6.
    \item A class is a valid class for an anchor if: 1) Objects of this class are unique in the current scene. 2) The number of objects of class in the current scene cannot exceed 6.
    \item Besides, an anchor can never belong to the same class as the target and, as such, its distractors.
\end{itemize}

Next, we elaborate on detailed rules for the generation of different spatial relationships:
\begin{itemize}
    \item \textbf{Allocentric:} For each anchor, we traverse all the valid target classes. For all the objects of a certain target class, we calculate the positional vector between the target and anchor objects. Combined with the anchor's own orientation vector, we determine whether the target is in front of, behind, to the left, or to the right of the anchor. Note that an allocentric language prompt will only be generated when there is only one object belonging to a certain target class in a certain direction of this anchor.
    \item \textbf{Support and Vertical Proximity:} For each anchor and target object, we first calculate the Intersection over Union (IoU) of the anchor and target in the XY plane. If the IoU exceeds a certain threshold, we determine whether the anchor and target can form a support relationship based on a pre-defined list of supporter and supportee categories. The positional relationship, above or below, is judged based on the heights in the Z-axis direction.
    \item \textbf{Horizontal Proximity:} For each anchor, we traverse all the valid target classes. For all the objects of a certain target class, we calculate their distances to the anchor object. From these, we select the farthest and nearest objects to construct a language prompt for each.
    \item \textbf{Between:} Unlike other types of prompts, this prompt requires two anchors. We determine whether the target is between two anchors based on their top view 2D bounding boxes. Generally speaking, the target should be in the same Z range for each of the two anchors and be away from every other distractors by a certain distance.
\end{itemize}

\begin{figure}
    \centering
    \includegraphics[width=1.0\linewidth]{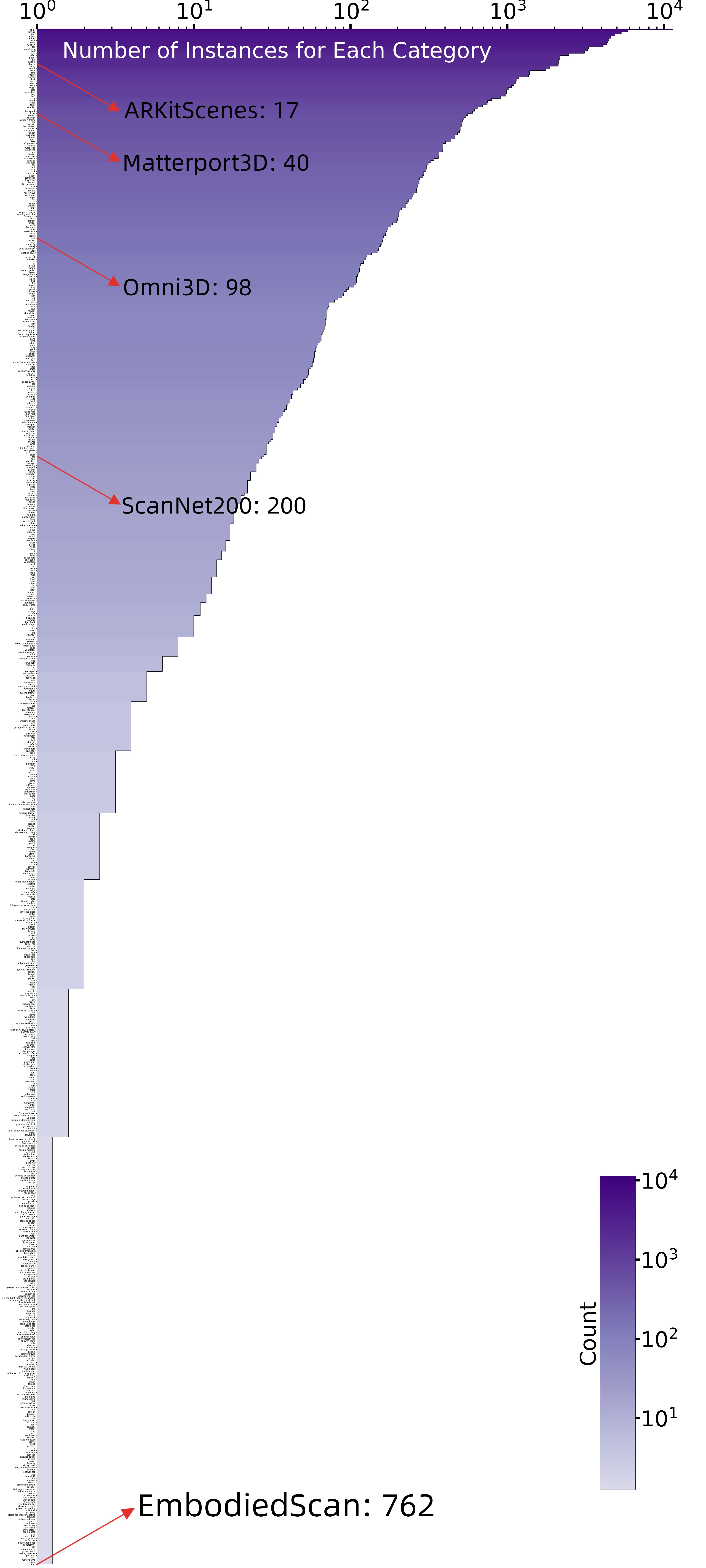}
    \caption{Complete instance distribution of EmbodiedScan.}
    \label{fig:supp-dist}
\vspace{-3ex}
\end{figure}

\subsection{Statistics}
\noindent\textbf{Complete Instance Statistics.}
We show the complete instance distribution in Fig.~\ref{fig:supp-dist} for reference. It can be observed that it turns out a long-tailed distribution as expected and shows obvious superiority over previous datasets regarding the number of categories and instances.

\noindent\textbf{Spatial Proximity Statistics.}
Here we first conduct a basic quantitative analysis of different categories of language prompts. We find that in the generated prompts, descriptions of Horizontal Proximity and Allocentric relationships accounted for the vast majority, while Vertical Proximity and Support only made up a small portion. This is consistent with the fact that most objects in the 3D scene are primarily distributed on the XY plane (horizontal direction).

\begin{table}
\scriptsize
\caption{Spatial proximity statistics.}
\vspace{-2ex}
    \centering
{%
\begin{tabular}{|c|c|c|c|c|c|}
\hline
Horizontal & Vertical & Support & Allocentric & Between & All \\
\hline
723477 & 16420 & 4812 & 216197 & 9135 & 970041 \\
\hline
\end{tabular}
}
\vspace{-4ex}
    \label{tab:proximity-stat}
\end{table}

In addition, we make further analysis of the prompt distribution and discover several reasonable statistic results:

1. For common objects such as tables, we find that in the generated Support prompts, the ten most frequent objects are
book, lamp, jacket, paper, plant, bottle, plate, box, telephone, and TV. By analyzing horizontal prompts, we find that nearby objects often include window, door, couch, cabinet, curtain, bin, and chair. This is actually consistent with our common sense, as tables are usually placed next to windows and accompanied by chairs for people to sit.

2. For smaller objects like books, we find that in the generated prompts, Support prompts account for the majority. Based on data analysis, books are usually placed on tables, stands, desks, cabinets, boxes, and dressers.

3. In the generated prompts, there are certain categories of objects usually appearing together within a scene. However, these objects are not common in our daily in-house lives, such as menu, cube, ridge, panel, sack, crate, and shovel.

4. We also discovered some fixed spatial relationships. Some objects often appear in pairs. For example, mirrors frequently appear above the sink, stool, cabinet, and socket in the generated prompts. On the other hand, pictures are often placed above the bed, couch, table, desk, toilet, and cabinet.

These findings are small but interesting. We believe that given such detailed instance annotations, further analyses for the object distributions can reflect some common sense regarding the daily object configurations. It would also provide useful guidance for AI-powered realistic 3D scene generation and design.

\begin{figure*}[t!]
     \centering
          \begin{subfigure}[b]{0.48\textwidth}
         \centering
         \includegraphics[width=\textwidth]{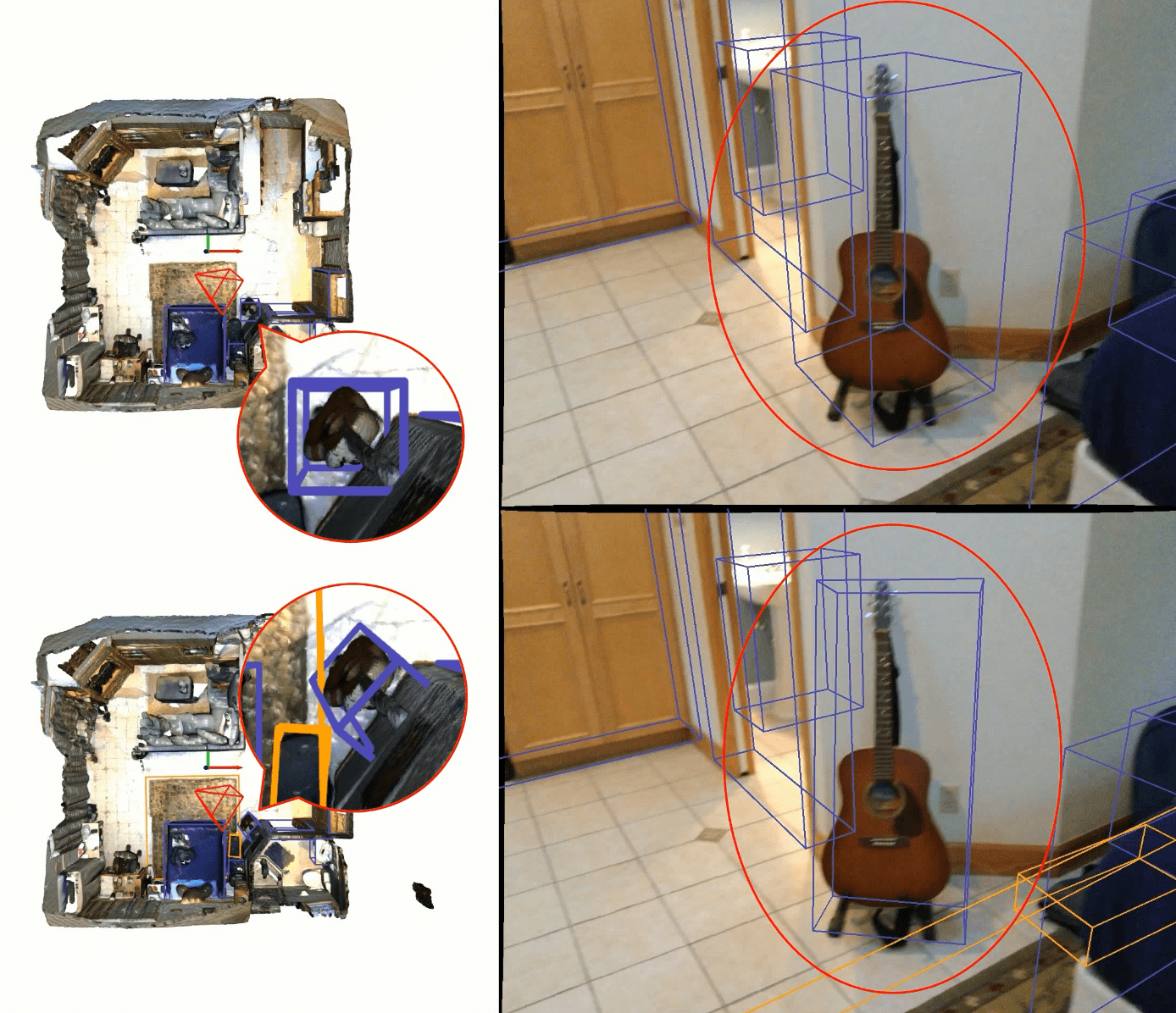}
         \caption{Box orientations.}
         \label{fig: supp-orientation} 
     \end{subfigure}
     \hfill
     \begin{subfigure}[b]{0.48\textwidth}
        \centering
        \includegraphics[width=\textwidth]{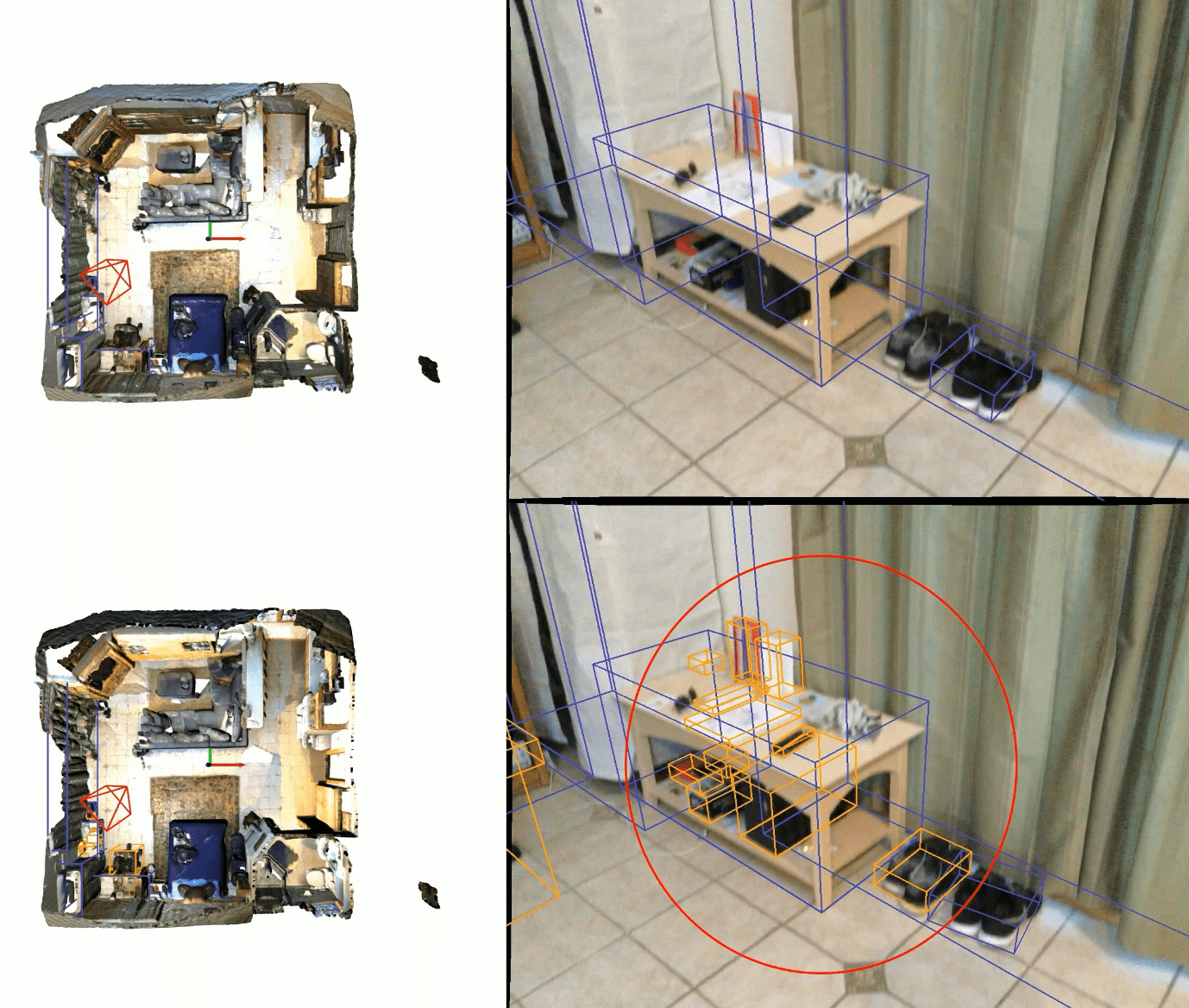}
        \caption{Small objects (new boxes marked in yellow).}
        \label{fig: supp-small-boxes} 
     \end{subfigure}
     \caption{Comparison of previous (top) and our annotations (bottom).} 
     \label{fig: data-samples}
     \vspace{-2.5ex}
\end{figure*}



\subsection{Data Examples}
\noindent\textbf{Oriented 3D Boxes.} We show the comparison of previous and current annotations in Fig.~\ref{fig: data-samples}. Here, we highlight the difference in box orientations and new annotated small objects in the provided two examples.

\noindent\textbf{Complicated Language Prompts.}
Due to the fact that the language descriptions generated in the new benchmark contain more object categories, it is easier to generate prompts with ambiguity.
Although we tried our best to ensure that the generated prompts have unique references through various restrictions during the generation procedure, this phenomenon still exists in our later manual inspection.
Therefore, we randomly concatenate multiple generated language prompts belonging to the same object to generate complicated language prompts for auxiliary training. For example, for one object with three generated language prompts: ``find the monitor that is closer to the door", ``the monitor that is farthest from the windowsill" and ``the monitor that is near the fan", we combine them to obtain the complicated language prompt: ``find the monitor that is closer to the door, and it is farthest from the windowsill and near the fan."

\subsection{Clarifications}
\noindent\textbf{Dataset Comparison.}
First, we clarify several details in Tab. 1 of the main paper. For a fair and clear comparison, we modify some raw statistics of those datasets to make them consistent with the criterion of our dataset. For example, we do not show the number of objects for monocular datasets such as SUN RGB-D and Hypersim because these numbers can be inaccurate due to the potential repetitive counting of objects across different frames. We normalize the number of images from ScanNet by dividing its frame numbers by 10 to keep its sampling frequency consistent with ours. For categories, we only list the number of categories used for previous 3D detection instead of also involving that for 3D instance segmentation to highlight the much larger vocabulary of our 3D box annotations.

In addition, the number of language prompts shown in the Table is from ScanRefer~\cite{scanrefer}. There are also other annotations built upon ScanNet and we supplement these works here. ReferIt3D~\cite{referit3d} is another work concurrent to ScanRefer but focuses on fine-grained 3D object identification, providing 120k prompts including spatial reference (SR3D) and natural reference (NR3D).
ScanQA~\cite{scanqa} targets the question-answering problem in 3D scenes and offers 41k question-answering pairs for ScanNet.
SQA3D~\cite{sqa3d} further highlights the role of ``situation" in this problem, resulting in 21k descriptions of 6.8k unique situations and 35k questions.
All these works are built upon only ScanNet and thus have limited scene diversity.
Recently, to collect large-scale 3D-text pairs for pre-training, 3D-VisTA~\cite{3d-vista} generates 278k scene descriptions from existing 3D Vision-Language tasks, templates, and GPT-3 from ScanNet and 3R-Scan as ScanScribe. It also randomly replaces objects from Objaverse with the same class to enhance the scene and object diversity, but at the same time, may yield a little domain gap between the generated and real-scanned raw data.
In contrast, thanks to our comprehensive annotations for objects, regarding both categories and object poses, and more diverse scans, our preliminary version has a much larger scale in the language descriptions, scaling up the number to about 1M. Furthermore, due to the complicated scenes and object distributions, the task also becomes more challenging. We will continue to improve the existing language annotations and add more content from other aspects for holistic 3D scene understanding.

\noindent\textbf{Test Set.} We respect the copyright and license of all the source datasets and only include the test set statistics for scans and images in the main paper. For the annotations of the test set, we connect with the official hosts and will consider making a more complete version for future benchmarks and challenges. Other related issues will also be addressed by clear communication and collaboration with the official hosts before the data and models are released in the future.

\begin{table*}[t]
\scriptsize
\setlength{\tabcolsep}{3.5pt}
\caption{Continuous and multi-view 3D detection results per category.}
\vspace{-2ex}
\centering
{%
    \begin{tabular}{c|c|cccccccccccccccccc}
    \hline
    Methods & mAP$_{25}$  & chair & picture & door & pillow & cabinet & table & book & window & box & shelf & plant & bin & curtain & bottle & lamp & couch & towel & sink\\
    \hline
    Camera-Only & 12.80 & 72.39 & 1.30 & 26.63 & 25.07 & 21.08 & 55.03 & 2.81 & 6.16 & 8.25 & 25.37 & 28.20 & 42.66 & 8.84 & 0.01 & 12.87 & 75.90 & 0.46 & 26.93 \\
    Depth-Only & 17.16 & 80.68 & 6.88 & 29.77 & 40.89 & 22.25 & 67.83 & 1.28 & 31.61 & 10.02 & 45.93 & 29.90 & 28.52 & 6.01 & 2.55 & 31.85 & 70.14 & 45.08 & 67.16 \\
    Multi-Modality & \textbf{19.07} & 80.73 & 15.59 & 35.45 & 51.46 & 25.22 & 62.14 & 4.56 & 24.82 & 10.45 & 45.98 & 48.32 & 30.07 & 14.60 & 2.02 & 34.62 & 78.12 & 40.72 & 65.68\\
    \hline\hline
    ImVoxelNet~\cite{imvoxelnet} & 6.15 & 69.76 & 0.41 & 10.79 & 15.38 & 14.43 & 45.63 & 1.77 & 4.82 & 5.50 & 13.36 & 12.96 & 29.96 & 9.79 & 0.01 & 15.44 & 54.51 & 1.94 & 22.76 \\
    VoteNet~\cite{votenet} & 3.20 & 64.92 & 0.00 & 0.01 & 3.30 & 3.24 & 30.42 & 0.12 & 0.05 & 0.91 & 0.02 & 1.84 & 17.17 & 0.39 & 0.00 & 5.55 & 33.37 & 0.11 & 10.88 \\
    \hline
    FCAF3D~\cite{fcaf3d} & 9.07 & 86.98 & 2.42 & 9.01 & 44.54 & 21.03 & 54.20 & 15.02 & 10.71 & 7.13 & 24.65 & 23.22 & 56.93 & 17.86 & 0.47 & 27.11 & 63.56 & 11.80 & 63.74 \\
    +our decoder & 14.80 & 90.30 & 17.01 & 42.82 & 49.22 & 36.01 & 67.20 & 20.94 & 30.26 & 9.83 & 41.93 & 30.40 & 70.51 & 39.44 & 1.13 & 35.33 & 76.99 & 36.01 & 72.58 \\
    +painting & 15.10 & 90.79 & 20.25 & 45.70 & 52.30 & 36.98 & 67.42 & 18.55 & 31.23 & 11.30 & 40.98 & 33.14 & 70.28 & 38.27 & 0.91 & 34.50 & 73.70 & 30.45 & 73.43 \\
    \hline
    Ours & \textbf{16.85} & 88.81 & 19.57 & 42.36 & 54.65 & 38.78 & 67.12 & 20.59 & 33.69 & 12.92 & 40.97 & 35.48 & 71.18 & 43.85 & 1.52 & 37.36 & 77.65 & 31.74 & 72.92\\
    \hline
    \end{tabular}%
}
\vspace{-2ex}
\label{tab:mvdet3d-category}
\end{table*}

\begin{table*}[t]
\footnotesize
\setlength{\tabcolsep}{3.5pt}
\caption{Monocular 3D detection results per category.}
\vspace{-2ex}
\centering
{%
    \begin{tabular}{c|c|cccccccccccccccccccc}
    \hline
    Methods & mAP$_{25}$  & chair & pillow & cabinet & table & lamp & couch & desk & stand & bed & backpack \\
    \hline
    FCOS3D~\cite{FCOS3D} & 8.93 & 27.15 & 2.23 & 1.14 & 6.21 & 1.92 & 9.47 & 12.09 & 11.13 & 18.38 & 5.52  \\
    ImVoxelNet~\cite{imvoxelnet} & 18.95 & 46.70 & 5.93 & 4.63 & 18.10 & 6.58 & 20.39 & 24.78 & 19.58 & 41.51 & 14.64 \\
    VoteNet~\cite{votenet} & 14.30 & 54.00 & 1.65 & 2.41 & 19.53 & 3.55 & 21.80 & 19.13 & 4.89 & 45.58 & 4.21 \\
    ImVoteNet~\cite{imvotenet} & 19.63 & 56.72 & 2.10 & 2.88 & 29.00 & 10.01 & 27.77 & 23.13 & 12.68 & 56.94 & 10.93 \\
    \hline
    FCAF3D~\cite{fcaf3d} & 25.70 & 65.91 & 23.19 & 6.47 & 26.64 & 17.87 & 22.50 & 31.64 & 25.03 & 53.68 & 28.24 \\
    +our decoder & 28.16 & 63.85 & 28.68 & 6.62 & 32.34 & 14.19 & 31.61 & 30.81 & 27.27 & 60.03 & 32.43 \\
    +painting & 30.19 & 66.39 & 28.28 & 7.41 & 33.66 & 18.23 & 32.24 & 35.64 & 29.69 & 60.04 & 37.92 \\
    \hline
    Ours & \textbf{34.28} & 69.47 & 31.64 & 10.01 & 37.29 & 19.73 & 31.67 & 39.07 & 32.01 & 63.27 & 37.89 \\
    \hline\hline
    Methods & mAP$_{25}$ & bathtub & ottoman & dresser & bin & toilet & refri. & stove & microwave & monitor & computer \\
    \hline
    FCOS3D~\cite{FCOS3D} & 8.93 & 6.31 & 1.38 & 4.54 & 10.23 & 40.51 & 6.92 & 4.03 & 5.60 & 3.25 & 0.57 \\
    ImVoxelNet~\cite{imvoxelnet} & 18.95 & 10.14 & 9.63 & 9.98 & 17.82 & 65.70 & 18.11 & 15.92 & 14.93 & 8.26 & 5.80 \\
    VoteNet~\cite{votenet} & 14.30 & 13.49 & 7.60 & 0.53 & 14.72 & 68.16 & 0.96 & 1.35 & 0.16 & 1.26 & 1.08 \\
    ImVoteNet~\cite{imvotenet} & 19.63 & 37.56 & 9.14 & 1.87 & 21.96 & 74.08 & 1.21 & 9.50 & 2.12 & 2.24 & 0.66 \\
    \hline
    FCAF3D~\cite{fcaf3d} & 25.70 & 26.38 & 15.76 & 4.35 & 34.93 & 71.90 & 13.88 & 4.29 & 9.95 & 21.57 & 9.79 \\
    +our decoder & 28.16 & 38.17 & 21.85 & 7.28 & 38.96 & 75.57 & 16.25 & 7.78 & 10.31 & 6.13 & 13.04 \\
    +painting & 30.19 & 41.31 & 20.23 & 7.16 & 42.86 & 77.59 & 16.12 & 9.56 & 10.76 & 14.04 & 14.68 \\
    \hline
    Ours & \textbf{34.28} & 50.63 & 25.59 & 9.54 & 45.17 & 80.39 & 24.44 & 14.53 & 19.96 & 19.77 & 23.65 \\
    \hline
    \end{tabular}%
}
\vspace{-2ex}
\label{tab:mono3d-category}
\end{table*}

\section{Supplementary Results}
Due to the space limitation, we only list the main benchmark results and key ablation studies to demonstrate the value of our dataset and the efficacy of our baseline. Here, we further show more details about these results, supplement more ablation studies, and visualize the predictions qualitatively both on our dataset and in the real world.

\subsection{Detailed Benchmark Results}
\noindent\textbf{3D Detection Results Per Category.}
First, we show the detailed continuous and multi-view 3D detection performance in Tab.~\ref{tab:mvdet3d-category} for categories that are common in the real world and annotations. We can see that although these categories have a large number of annotations, there are still some that seem challenging for current models, such as pictures and bottles. In addition, it can be observed that the improvement brought by a better decoder for orientation estimation (+our decoder) is mainly focused on those objects that have significant differences between length and width, such as pictures, doors, windows, shelves, towels, \emph{etc.} It is a reasonable phenomenon and reveals the importance of orientation estimation in this setting. Finally, because we do not list the 20 categories in the monocular 3D detection benchmark, we show the complete results in Tab.~\ref{tab:mono3d-category}.

\subsection{Supplementary Ablation Studies}
\begin{figure*}[t!]
     \centering
     \begin{subfigure}[b]{0.31\textwidth}
         \centering
         \includegraphics[width=\textwidth]{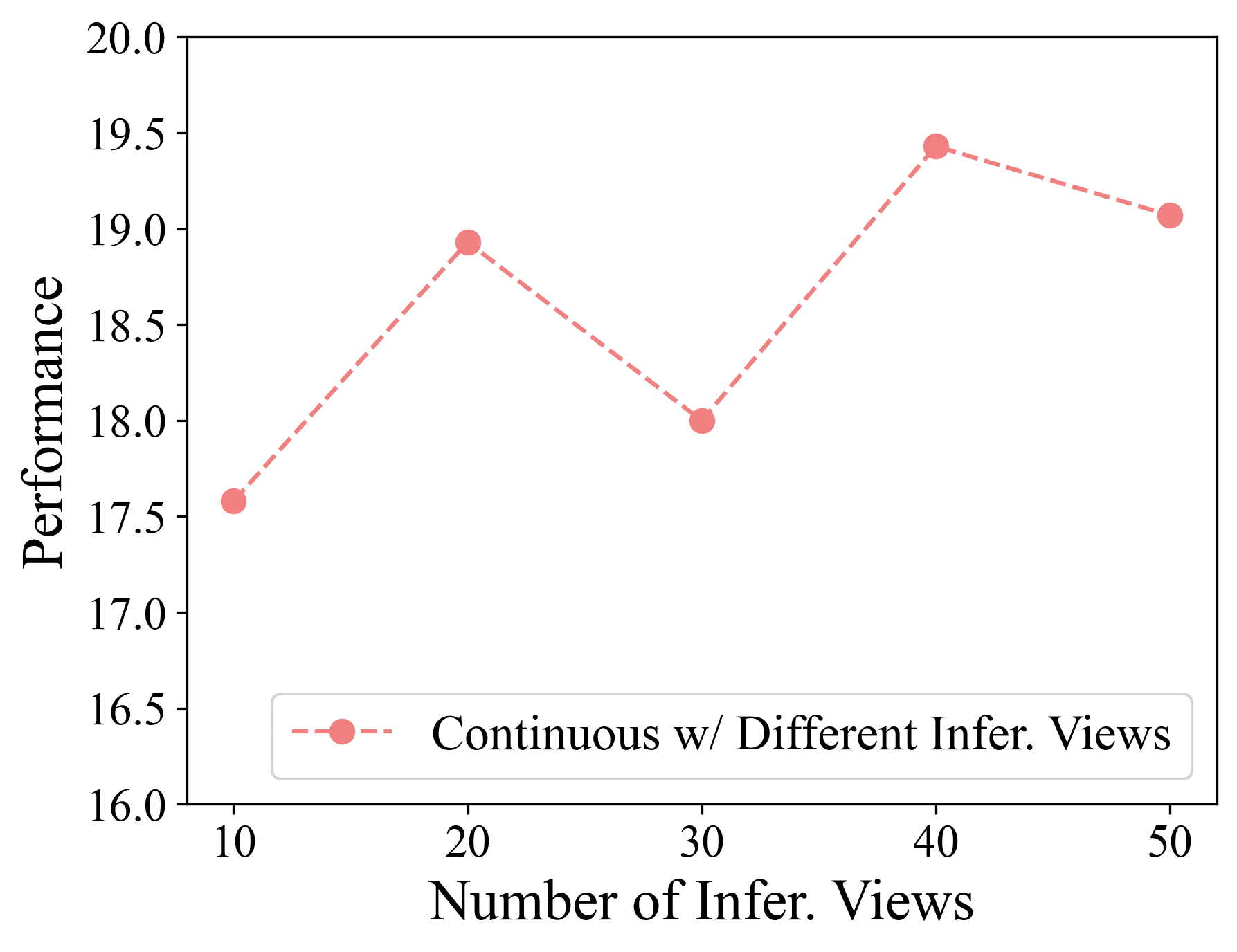}
         \caption{Continuous 3D detection with different numbers of inference views.}
         \label{fig: scale-views-continuous} 
     \end{subfigure}
     \hfill
     \begin{subfigure}[b]{0.31\textwidth}
        \centering
        \includegraphics[width=\textwidth]{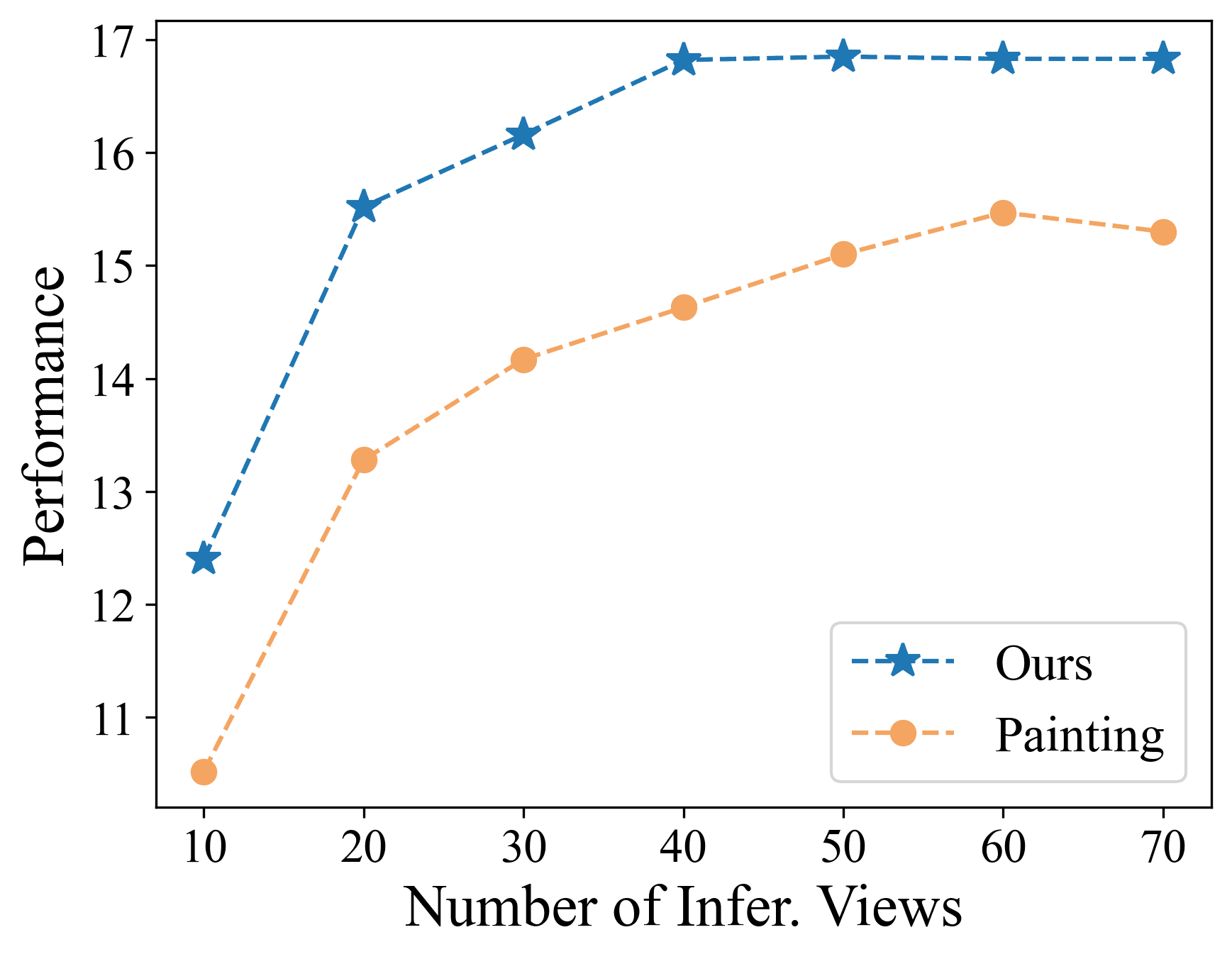}
        \caption{Multi-view 3D detection with different numbers of inference views.}
        \label{fig: scale-views-mv} 
     \end{subfigure}
     \hfill
     \begin{subfigure}[b]{0.31\textwidth}
         \centering
         \includegraphics[width=\textwidth]{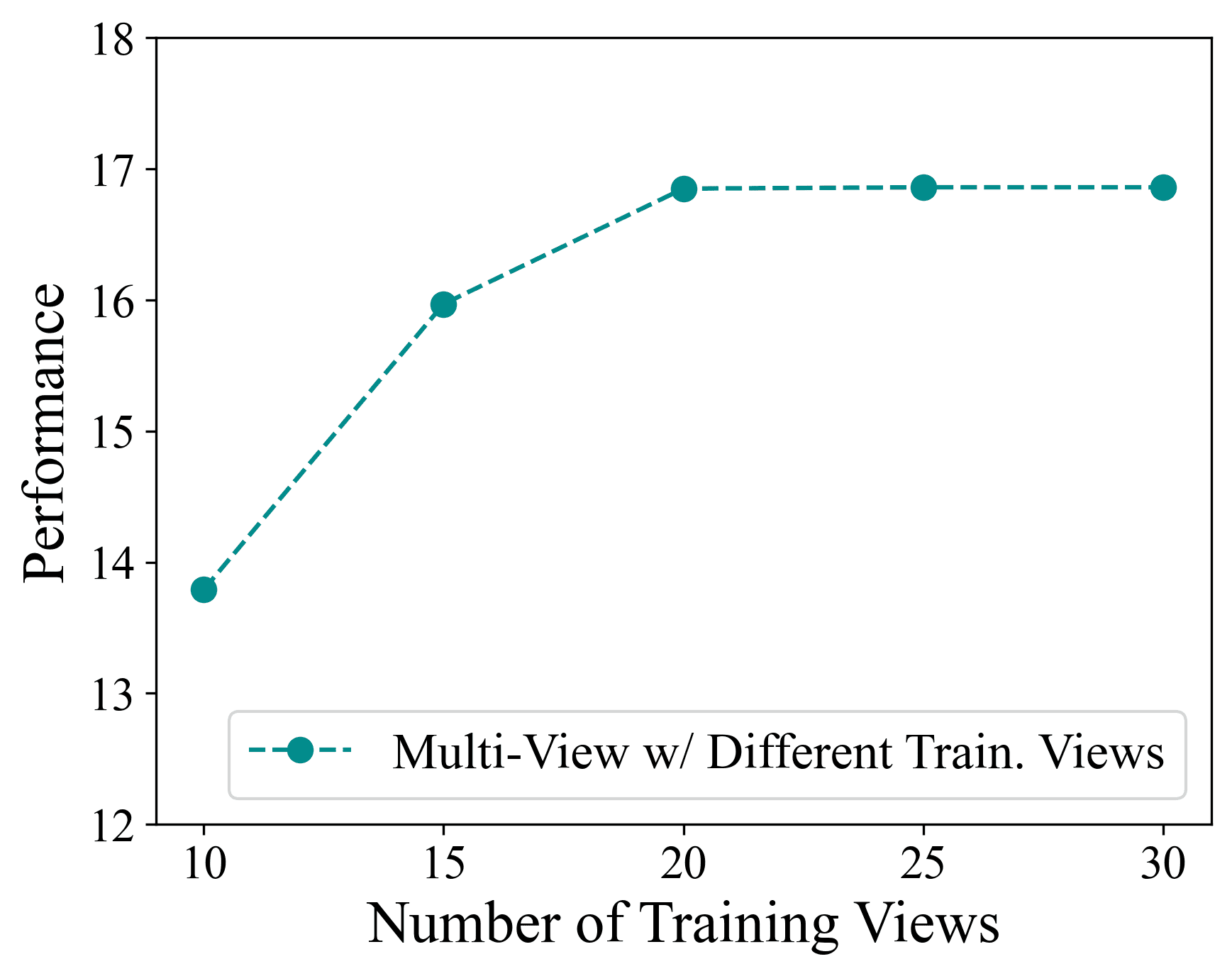}
         \caption{Multi-view 3D detection with different numbers of training views.}
         \label{fig: scale-views-train} 
     \end{subfigure}
     \caption{Performance changed with the number of training and inference views.} 
     \label{fig: scale-views}
\end{figure*}

\noindent\textbf{Number of Views.} 
As mentioned in the main paper, our trained baseline is applicable to any number of views during training and inference. Here, we show the ablation study regarding the number of views used for training and inference in Fig.~\ref{fig: scale-views}. Specifically, taking 3D detection as the example, we change the inference views for continuous and multi-view settings and record the performance change in Fig.~\ref{fig: scale-views-continuous} and \ref{fig: scale-views-mv}. We can see that it has a relatively minor influence on the continuous setting because the ground truth also changes as the visible instances become fewer when reducing the number of inference views. For multi-view experiments, it affects the performance only when the number of views is too small (\emph{e.g.}, $<20$), but it is more robust than the simple painting baseline, potentially benefiting from the stronger multi-modality fusion. Finally, similar to inference, it is also better to use more views for training but would saturate when using more than 20 views (Fig.~\ref{fig: scale-views-train}). Therefore, setting the number of training views to 20 is a good trade-off between training costs and performance.

\noindent\textbf{Dense Fusion.}
Here, we provide more comparison results with other RGB-D baselines for dense occupancy prediction. Taking the multi-view occupancy prediction as an example, we re-implement the painting baseline as the detection experiments and observe a much lower performance, as shown in Tab.~\ref{tab:dense-fusion}. We conjecture it is because the painting loses much more dense information with such sparse feature extraction, resulting in such a larger gap from our dense fusion method.
As for the alternative choices for dense fusion, we first attempt to voxelize the space with 0.16m voxels and use a MinkUNet to produce the sparse voxel feature for subsequent fusion and dense prediction. It turns out that fine-grained partition is necessary at the beginning. Besides, if removing the FPN to make the 2D feature extraction more lightweight, we cannot obtain the final competitive performance either.

\begin{table*}[t]
\scriptsize
\setlength{\tabcolsep}{3.5pt}
\caption{Ablation studies for dense fusion.}
\vspace{-2ex}
\centering
{%
    \begin{tabular}{c|c|c|cccccccccccccccc}
    \hline
    Methods  & Input & mIOU  & empty & floor &  wall & chair & cabinet & door & table & couch & shelf & window & bed & curtain & refri. & plant & stairs & toilet \\
    \hline
    Painting & RGB-D & 20.33 & 77.45 & 70.49 & 57.58 & 55.11 & 31.51 & 22.29 & 55.61 & 47.75 & 40.27 & 28.16 & 50.32 & 41.52 & 19.14 & 19.66 & 13.64 & 45.43 \\
    MinkUNet & RGB-D & 24.53 & 71.10 & 63.41 & 56.19 & 49.99 & 35.05 & 37.16 & 50.96 & 46.87 & 38.30 & 33.56 & 54.97 & 41.49 & 30.20 & 35.89 & 12.77 & 64.56 \\
    MinkResNet (w/o FPN) & RGB-D & 21.16 & 77.45 & 70.81 & 57.09 & 55.44 & 30.99 & 23.49 & 55.31 & 49.64 & 40.59 & 30.65 & 48.58 & 38.96 & 19.10 & 22.37 & 6.77 & 58.57 \\
    MinkResNet (Ours) & RGB-D & \textbf{27.65} & 77.57 & 71.04 & 62.12 & 57.30 & 38.31 & 41.09 & 56.79 & 50.72 & 46.06 & 38.75 & 56.24 & 46.38 & 29.47 & 40.52 & 17.44 & 68.95 \\
    \hline
    \end{tabular}%
}
\vspace{-2ex}
\label{tab:dense-fusion}
\end{table*}

\noindent\textbf{Sparse Fusion.}
However, the FPN is not necessary for sparse fusion, especially considering the optimization problem encountered in the 3D detection baseline. Furthermore, except for the unstable training problem mentioned in the main paper, our baseline is also much more computationally efficient than the alternative implementations, which keep the FPN or paint the points with image features. Our final baseline costs only $\sim$25G of memory with the reduction of 2D feature channels and removing the FPN, compared to $\sim$59G of memory used in other approaches.

\noindent\textbf{Sparse Decoder.}
The basic comparison between the simple L1 loss for Euler angles and our final decoder has been shown in the main paper. Here, we elaborate more on the ablation results of several design details in Tab.~\ref{tab:sparse-decoder}. We first compare the results of different combination methods for the corner losses and see that the weighted disentangled loss shows the best performance, compared to ``w/o decouple", ``simple summation", and ``average". In addition, taking the size of boxes into consideration and normalizing the corner losses by their sizes cannot bring improvement in the final performance.

Apart from the corner loss, another straightforward approach is to implement a pseudo-3D-IoU loss for these methods. Specifically, since the computation of 3D IoU among 9-DoF boxes is still heavy and non-differentiable, we approximate the 3D boxes as 7-DoF ones with only the yaw part in the axis-aligned coordinate system. This ``hack" method shows outstanding performance, especially in mAP$_{50}$, and can be further enhanced by combining the corner loss. Therefore, IoU-based loss is a design more faithful to the final metric and worthy of further study for the general 9-DoF case.

\begin{table}
\scriptsize
\caption{Ablation studies of designs for sparse decoder.}
\vspace{-4.5ex}
\begin{center}
{
    \begin{tabular}{c|cc|ccc}
    \hline
     Method & mAP$_{25}$ & mAP$_{50}$ & Head$_{25}$ & Common$_{25}$ & Tail$_{25}$ \\
    \hline
    w/o Decouple & 20.14 & 11.89 & 30.71 & 14.06 & 6.16 \\
    Decouple (sum.) & 18.03 & 9.98 & 27.16 & 12.85 & 5.86 \\
    Decouple (avg.) & 21.50 & 11.30 & 32.13 & 17.62 & 3.71 \\
    +Norm by 3D Size & 20.67 & 11.39 & 30.68 & 16.20 & 5.28 \\
    Decouple (weigh.) & \textbf{21.70} & 12.53 & 31.77 & 16.89 & 6.77 \\
    \hline
    7-DoF IoU Loss & 21.51 & 14.43 & 32.21 & 16.03 & 6.22 \\
    + Corner Loss & \textbf{22.13} & 13.95 & 32.60 & 16.82 & 7.08 \\
    \hline
    \end{tabular}
}
\vspace{-8ex}
\end{center}
\label{tab:sparse-decoder}
\end{table}

\noindent\textbf{Performance with Different Training Data.}
During the procedure of scaling up data and annotations, we also test the model's performance on ScanNet and our final validation set. We can find a performance improvement that seems to be linear with respect to the number of scans (1.5k-3k-5k scans from ScanNet to EmbodiedScan), especially for objects with plenty of annotations (``Head" categories). We would continue collecting the RGB-D scan data and annotations to further push the model's performance to a higher level, towards the usage in practice and real-world embodied AI.

\begin{table}
\scriptsize
\caption{Performance with different training data.}
\vspace{-4.5ex}
\begin{center}
{
    \begin{tabular}{c|c|c|ccc}
    \hline
     Train & Val & Overall & Head & Common & Tail \\
    \hline
    ScanNet & ScanNet & 20.28 & 29.81 & 15.57 & 6.40 \\
    +3RScan & ScanNet & 21.41 & 31.61 & 17.07 & 5.35\\
    +Matterport3D & ScanNet & \textbf{23.02} & 33.82 & 18.09 & 6.57 \\
    \hline
    ScanNet & EmbodiedScan & 10.92 & 21.10 & 8.06 & 1.78 \\
    +3RScan & EmbodiedScan & 13.91 & 25.25 & 10.69 & 3.76 \\
    +Matterport3D & EmbodiedScan & \textbf{16.85} & 28.65 & 12.83 & 7.09 \\
    \hline
    \end{tabular}
}
\vspace{-8ex}
\end{center}
\label{tab:different_training_data}
\end{table}

\begin{figure*}
    \centering
    \includegraphics[width=1.0\linewidth]{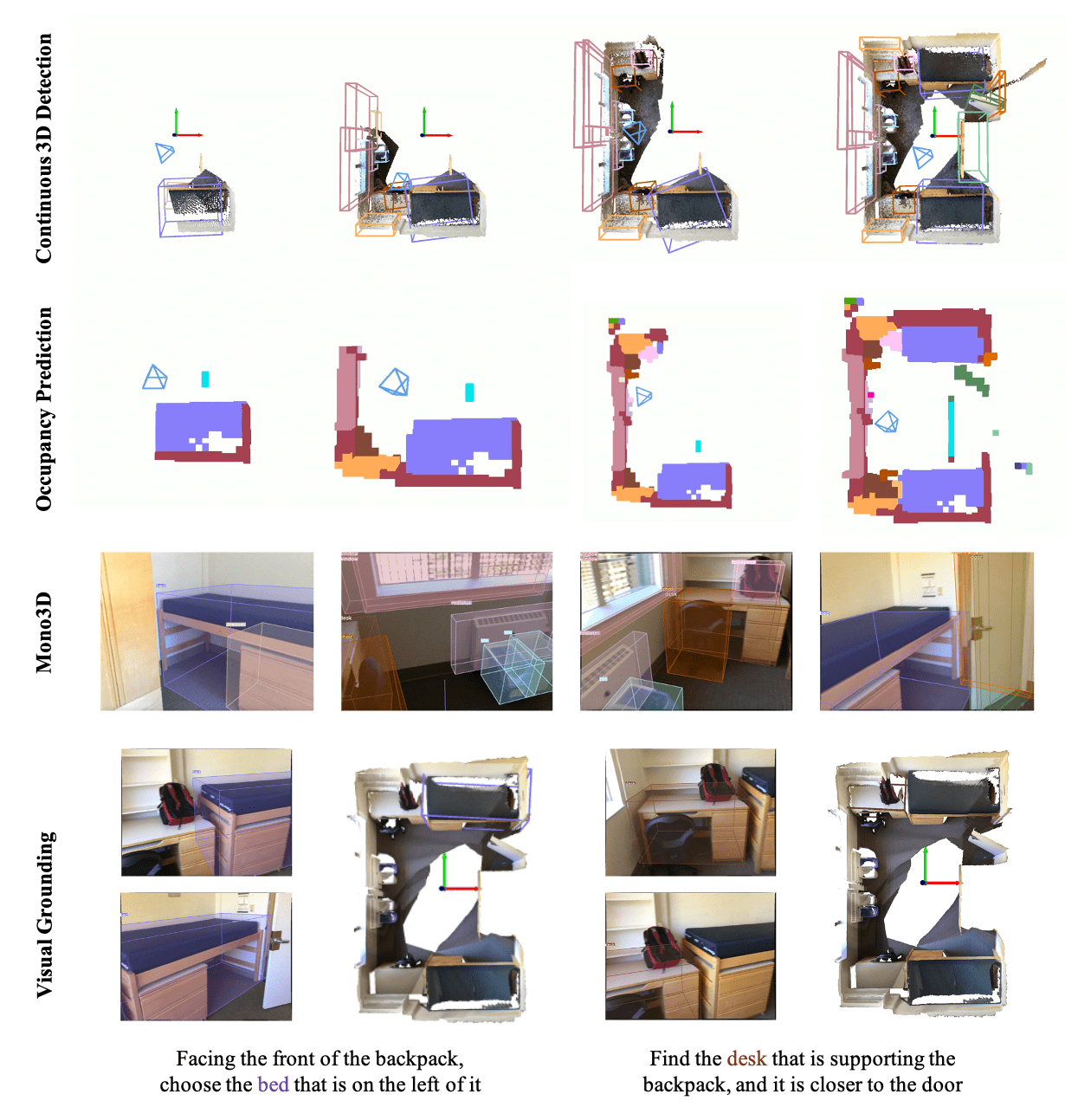}
    \caption{Qualitative results of different tasks on EmbodiedScan.}
    \label{fig:qualitative}
\end{figure*}

\subsection{Qualitative Results}
We visualize the prediction results on EmbodiedScan in Fig.~\ref{fig:qualitative}. From top to bottom, we plot the predictions of continuous 3D detection and occupancy prediction, monocular 3D detection, and multi-view 3D visual grounding. From this visualization, we can have a feeling about different perception output formats and how our models perform on our dataset. We can observe that the continuous perception can keep most previous predictions and fix some of them with the exploration. In addition, the localization of 3D visual grounding can be more accurate than the classical 3D detection for the target object considering it only needs to predict a single bounding box.

\subsection{In-the-Wild Test Demo}
Finally, we test our trained model in the wild. It shows decent performance in our test cases without cherry-picking, even with a different RGB-D sensor (Kinect) in different environments potentially having significant domain gaps from the training data. We visualize the prediction results in our demo video.

\section{Demo Video}
To provide a summary of our paper and key contributions, we made a short demo video, covering the overview of our dataset and methodology, more visualization of the annotation tool and perception results, to give a more intuitive understanding of our work. Please see the video with more details on our project page \href{http://tai-wang.github.io/embodiedscan}{http://tai-wang.github.io/embodiedscan}.

{
    \small
    \bibliographystyle{ieeenat_fullname}
    \bibliography{egbib}

\begin{thebibliography}{68}
\providecommand{\natexlab}[1]{#1}
\providecommand{\url}[1]{\texttt{#1}}
\expandafter\ifx\csname urlstyle\endcsname\relax
  \providecommand{\doi}[1]{doi: #1}\else
  \providecommand{\doi}{doi: \begingroup \urlstyle{rm}\Url}\fi

\bibitem[Achlioptas et~al.(2020)Achlioptas, Abdelreheem, Xia, Elhoseiny, and Guibas]{referit3d}
Panos Achlioptas, Ahmed Abdelreheem, Fei Xia, Mohamed Elhoseiny, and Leonidas Guibas.
\newblock Referit3d: Neural listeners for fine-grained 3d object identification in real-world scenes.
\newblock In \emph{European conference on computer vision}, 2020.

\bibitem[Azuma et~al.(2022)Azuma, Miyanishi, Kurita, and Kawanabe]{scanqa}
Daichi Azuma, Taiki Miyanishi, Shuhei Kurita, and Motoaki Kawanabe.
\newblock Scanqa: 3d question answering for spatial scene understanding.
\newblock In \emph{proceedings of the IEEE/CVF conference on computer vision and pattern recognition}, 2022.

\bibitem[Baruch et~al.(2021)Baruch, Chen, Dehghan, Dimry, Feigin, Fu, Gebauer, Joffe, Kurz, Schwartz, and Shulman]{arkitscenes}
Gilad Baruch, Zhuoyuan Chen, Afshin Dehghan, Tal Dimry, Yuri Feigin, Peter Fu, Thomas Gebauer, Brandon Joffe, Daniel Kurz, Arik Schwartz, and Elad Shulman.
\newblock {ARK}itscenes - a diverse real-world dataset for 3d indoor scene understanding using mobile {RGB}-d data.
\newblock In \emph{Thirty-fifth Conference on Neural Information Processing Systems Datasets and Benchmarks Track (Round 1)}, 2021.

\bibitem[Brazil et~al.(2023)Brazil, Kumar, Straub, Ravi, Johnson, and Gkioxari]{omni3d}
Garrick Brazil, Abhinav Kumar, Julian Straub, Nikhila Ravi, Justin Johnson, and Georgia Gkioxari.
\newblock {Omni3D}: A large benchmark and model for {3D} object detection in the wild.
\newblock In \emph{Proceedings of the IEEE Conference on Computer Vision and Pattern Recognition}, 2023.

\bibitem[Caesar et~al.(2019)Caesar, Bankiti, Lang, Vora, Liong, Xu, Krishnan, Pan, Baldan, and Beijbom]{nuScenes}
Holger Caesar, Varun Bankiti, Alex~H. Lang, Sourabh Vora, Venice~Erin Liong, Qiang Xu, Anush Krishnan, Yu Pan, Giancarlo Baldan, and Oscar Beijbom.
\newblock nuscenes: A multimodal dataset for autonomous driving.
\newblock \emph{CoRR}, abs/1903.11027, 2019.

\bibitem[Cao and de~Charette(2022)]{monoscene}
Anh-Quan Cao and Raoul de Charette.
\newblock Monoscene: Monocular 3d semantic scene completion.
\newblock In \emph{Proceedings of the IEEE Conference on Computer Vision and Pattern Recognition}, 2022.

\bibitem[Chang et~al.(2017)Chang, Dai, Funkhouser, Halber, Niessner, Savva, Song, Zeng, and Zhang]{Matterport3D}
Angel Chang, Angela Dai, Thomas Funkhouser, Maciej Halber, Matthias Niessner, Manolis Savva, Shuran Song, Andy Zeng, and Yinda Zhang.
\newblock {Matterport3D}: Learning from {RGB-D} data in indoor environments.
\newblock \emph{International Conference on 3D Vision (3DV)}, 2017.

\bibitem[Chang et~al.(2019)Chang, Lambert, Sangkloy, Singh, Bak, Hartnett, Wang, Carr, Lucey, Ramanan, et~al.]{argoverse}
Ming-Fang Chang, John Lambert, Patsorn Sangkloy, Jagjeet Singh, Slawomir Bak, Andrew Hartnett, De Wang, Peter Carr, Simon Lucey, Deva Ramanan, et~al.
\newblock Argoverse: 3d tracking and forecasting with rich maps.
\newblock In \emph{Proceedings of the IEEE/CVF conference on computer vision and pattern recognition}, 2019.

\bibitem[Chen et~al.(2020)Chen, Chang, and Nie{\ss}ner]{scanrefer}
Dave~Zhenyu Chen, Angel~X Chang, and Matthias Nie{\ss}ner.
\newblock Scanrefer: 3d object localization in rgb-d scans using natural language.
\newblock In \emph{European conference on computer vision}, 2020.

\bibitem[Chen et~al.(2021)Chen, Gholami, Nie{\ss}ner, and Chang]{scan2cap}
Zhenyu Chen, Ali Gholami, Matthias Nie{\ss}ner, and Angel~X Chang.
\newblock Scan2cap: Context-aware dense captioning in rgb-d scans.
\newblock In \emph{Proceedings of the IEEE/CVF conference on computer vision and pattern recognition}, pages 3193--3203, 2021.

\bibitem[Chen et~al.(2023)Chen, Hu, Chen, Nie{\ss}ner, and Chang]{Unit3D}
Zhenyu Chen, Ronghang Hu, Xinlei Chen, Matthias Nie{\ss}ner, and Angel~X Chang.
\newblock Unit3d: A unified transformer for 3d dense captioning and visual grounding.
\newblock In \emph{Proceedings of the IEEE/CVF International Conference on Computer Vision}, 2023.

\bibitem[Choy et~al.(2019)Choy, Gwak, and Savarese]{minkengine}
Christopher Choy, JunYoung Gwak, and Silvio Savarese.
\newblock 4d spatio-temporal convnets: Minkowski convolutional neural networks.
\newblock In \emph{Proceedings of the IEEE Conference on Computer Vision and Pattern Recognition}, 2019.

\bibitem[Contributors(2020)]{mmdet3d2020}
MMDetection3D Contributors.
\newblock {MMDetection3D: OpenMMLab} next-generation platform for general {3D} object detection.
\newblock \url{https://github.com/open-mmlab/mmdetection3d}, 2020.

\bibitem[Couprie et~al.(2013)Couprie, Farabet, Najman, and LeCun]{nyuv2}
Camille Couprie, Cl{\'e}ment Farabet, Laurent Najman, and Yann LeCun.
\newblock Indoor semantic segmentation using depth information.
\newblock \emph{arXiv preprint arXiv:1301.3572}, 2013.

\bibitem[Dai et~al.(2017)Dai, Chang, Savva, Halber, Funkhouser, and Nie{\ss}ner]{scannet}
Angela Dai, Angel~X Chang, Manolis Savva, Maciej Halber, Thomas Funkhouser, and Matthias Nie{\ss}ner.
\newblock Scannet: Richly-annotated 3d reconstructions of indoor scenes.
\newblock In \emph{Proceedings of the IEEE Conference on Computer Vision and Pattern Recognition}, 2017.

\bibitem[Devlin et~al.(2018)Devlin, Chang, Lee, and Toutanova]{bert}
Jacob Devlin, Ming-Wei Chang, Kenton Lee, and Kristina Toutanova.
\newblock Bert: Pre-training of deep bidirectional transformers for language understanding.
\newblock \emph{arXiv preprint arXiv:1810.04805}, 2018.

\bibitem[Ding et~al.(2023)Ding, Yang, Xue, Zhang, Bai, and Qi]{PLA}
Runyu Ding, Jihan Yang, Chuhui Xue, Wenqing Zhang, Song Bai, and Xiaojuan Qi.
\newblock Pla: Language-driven open-vocabulary 3d scene understanding.
\newblock In \emph{Proceedings of the IEEE/CVF Conference on Computer Vision and Pattern Recognition}, 2023.

\bibitem[Geiger et~al.(2012)Geiger, Lenz, and Urtasun]{KITTI}
Andreas Geiger, Philip Lenz, and Raquel Urtasun.
\newblock Are we ready for autonomous driving? the kitti vision benchmark suite.
\newblock In \emph{IEEE Conference on Computer Vision and Pattern Recognition}, 2012.

\bibitem[He et~al.(2016)He, Zhang, Ren, and Sun]{ResNet}
Kaiming He, Xiangyu Zhang, Shaoqing Ren, and Jian Sun.
\newblock Deep residual learning for image recognition.
\newblock In \emph{IEEE Conference on Computer Vision and Pattern Recognition}, 2016.

\bibitem[Huang et~al.(2022)Huang, Chen, Jia, and Wang]{mv3d-grounding}
Shijia Huang, Yilun Chen, Jiaya Jia, and Liwei Wang.
\newblock Multi-view transformer for 3d visual grounding.
\newblock In \emph{Proceedings of the IEEE/CVF Conference on Computer Vision and Pattern Recognition}, 2022.

\bibitem[Huang et~al.(2023)Huang, Zheng, Zhang, Zhou, and Lu]{tpvformer}
Yuanhui Huang, Wenzhao Zheng, Yunpeng Zhang, Jie Zhou, and Jiwen Lu.
\newblock Tri-perspective view for vision-based 3d semantic occupancy prediction.
\newblock In \emph{Proceedings of the IEEE/CVF conference on computer vision and pattern recognition}, 2023.

\bibitem[Jain et~al.(2022)Jain, Gkanatsios, Mediratta, and Fragkiadaki]{butd-detr}
Ayush Jain, Nikolaos Gkanatsios, Ishita Mediratta, and Katerina Fragkiadaki.
\newblock Bottom up top down detection transformers for language grounding in images and point clouds.
\newblock In \emph{European Conference on Computer Vision}, 2022.

\bibitem[{Khanna*} et~al.(2023){Khanna*}, {Mao*}, Jiang, Haresh, Shacklett, Batra, Clegg, Undersander, Chang, and Savva]{hssd}
Mukul {Khanna*}, Yongsen {Mao*}, Hanxiao Jiang, Sanjay Haresh, Brennan Shacklett, Dhruv Batra, Alexander Clegg, Eric Undersander, Angel~X. Chang, and Manolis Savva.
\newblock {Habitat Synthetic Scenes Dataset (HSSD-200): An Analysis of 3D Scene Scale and Realism Tradeoffs for ObjectGoal Navigation}.
\newblock \emph{arXiv preprint}, 2023.

\bibitem[Kirillov et~al.(2023)Kirillov, Mintun, Ravi, Mao, Rolland, Gustafson, Xiao, Whitehead, Berg, Lo, Doll{\'a}r, and Girshick]{SAM}
Alexander Kirillov, Eric Mintun, Nikhila Ravi, Hanzi Mao, Chloe Rolland, Laura Gustafson, Tete Xiao, Spencer Whitehead, Alexander~C. Berg, Wan-Yen Lo, Piotr Doll{\'a}r, and Ross Girshick.
\newblock Segment anything.
\newblock In \emph{Proceedings of the IEEE/CVF International Conference on Computer Vision}, 2023.

\bibitem[Lang et~al.(2019)Lang, Vora, Caesar, Zhou, Yang, and Beijbom]{PointPillars}
Alex~H. Lang, Sourabh Vora, Holger Caesar, Lubing Zhou, Jiong Yang, and Oscar Beijbom.
\newblock Pointpillars: Fast encoders for object detection from point clouds.
\newblock In \emph{IEEE Conference on Computer Vision and Pattern Recognition}, 2019.

\bibitem[Li et~al.(2020)Li, Wang, Li, Li, Wu, and Hao]{sustech}
E Li, Shuaijun Wang, Chengyang Li, Dachuan Li, Xiangbin Wu, and Qi Hao.
\newblock Sustech points: A portable 3d point cloud interactive annotation platform system.
\newblock In \emph{2020 IEEE Intelligent Vehicles Symposium (IV)}, 2020.

\bibitem[Li et~al.(2023)Li, Yu, Choy, Xiao, Alvarez, Fidler, Feng, and Anandkumar]{voxformer}
Yiming Li, Zhiding Yu, Christopher Choy, Chaowei Xiao, Jose~M Alvarez, Sanja Fidler, Chen Feng, and Anima Anandkumar.
\newblock Voxformer: Sparse voxel transformer for camera-based 3d semantic scene completion.
\newblock In \emph{Proceedings of the IEEE/CVF Conference on Computer Vision and Pattern Recognition}, 2023.

\bibitem[Liang et~al.(2022)Liang, Xie, Yu, Xia, Lin, Wang, Tang, Wang, and Tang]{liang2022bevfusion}
Tingting Liang, Hongwei Xie, Kaicheng Yu, Zhongyu Xia, Zhiwei Lin, Yongtao Wang, Tao Tang, Bing Wang, and Zhi Tang.
\newblock Bevfusion: A simple and robust lidar-camera fusion framework.
\newblock \emph{Advances in Neural Information Processing Systems}, 2022.

\bibitem[Lin et~al.(2017)Lin, Dollár, Girshick, He, Hariharan, and Belongie]{FPN}
Tsung-Yi Lin, Piotr Dollár, Ross Girshick, Kaiming He, Bharath Hariharan, and Serge Belongie.
\newblock Feature pyramid networks for object detection.
\newblock In \emph{IEEE Conference on Computer Vision and Pattern Recognition}, 2017.

\bibitem[Liu et~al.(2019)Liu, Ott, Goyal, Du, Joshi, Chen, Levy, Lewis, Zettlemoyer, and Stoyanov]{roberta}
Yinhan Liu, Myle Ott, Naman Goyal, Jingfei Du, Mandar Joshi, Danqi Chen, Omer Levy, Mike Lewis, Luke Zettlemoyer, and Veselin Stoyanov.
\newblock Roberta: A robustly optimized bert pretraining approach.
\newblock \emph{arXiv preprint arXiv:1907.11692}, 2019.

\bibitem[Liu et~al.(2020)Liu, Wu, and T{\'o}th]{smoke}
Zechen Liu, Zizhang Wu, and Roland T{\'o}th.
\newblock Smoke: Single-stage monocular 3d object detection via keypoint estimation.
\newblock In \emph{Proceedings of the IEEE/CVF Conference on Computer Vision and Pattern Recognition Workshops}, pages 996--997, 2020.

\bibitem[Liu et~al.(2021)Liu, Zhang, Cao, Hu, and Tong]{groupfree3d}
Ze Liu, Zheng Zhang, Yue Cao, Han Hu, and Xin Tong.
\newblock Group-free 3d object detection via transformers.
\newblock In \emph{Proceedings of the IEEE/CVF International Conference on Computer Vision}, 2021.

\bibitem[Liu et~al.(2023)Liu, Tang, Amini, Yang, Mao, Rus, and Han]{bevfusion}
Zhijian Liu, Haotian Tang, Alexander Amini, Xinyu Yang, Huizi Mao, Daniela~L Rus, and Song Han.
\newblock Bevfusion: Multi-task multi-sensor fusion with unified bird's-eye view representation.
\newblock In \emph{2023 IEEE International Conference on Robotics and Automation (ICRA)}, 2023.

\bibitem[Loshchilov and Hutter(2017)]{adamw}
Ilya Loshchilov and Frank Hutter.
\newblock Decoupled weight decay regularization.
\newblock \emph{arXiv preprint arXiv:1711.05101}, 2017.

\bibitem[Lu et~al.(2022)Lu, Xu, Wei, Xie, Tomizuka, Keutzer, and Zhang]{3detic}
Yuheng Lu, Chenfeng Xu, Xiaobao Wei, Xiaodong Xie, Masayoshi Tomizuka, Kurt Keutzer, and Shanghang Zhang.
\newblock Open-vocabulary 3d detection via image-level class and debiased cross-modal contrastive learning.
\newblock \emph{arXiv preprint arXiv:2207.01987}, 2022.

\bibitem[Ma et~al.(2022)Ma, Yong, Zheng, Li, Liang, Zhu, and Huang]{sqa3d}
Xiaojian Ma, Silong Yong, Zilong Zheng, Qing Li, Yitao Liang, Song-Chun Zhu, and Siyuan Huang.
\newblock Sqa3d: Situated question answering in 3d scenes.
\newblock \emph{arXiv preprint arXiv:2210.07474}, 2022.

\bibitem[Mao et~al.(2021)Mao, Niu, Jiang, Liang, Chen, Liang, Li, Ye, Zhang, Li, et~al.]{once}
Jiageng Mao, Minzhe Niu, Chenhan Jiang, Hanxue Liang, Jingheng Chen, Xiaodan Liang, Yamin Li, Chaoqiang Ye, Wei Zhang, Zhenguo Li, et~al.
\newblock One million scenes for autonomous driving: Once dataset.
\newblock \emph{arXiv preprint arXiv:2106.11037}, 2021.

\bibitem[Peng et~al.(2023)Peng, Genova, Jiang, Tagliasacchi, Pollefeys, Funkhouser, et~al.]{openscene}
Songyou Peng, Kyle Genova, Chiyu Jiang, Andrea Tagliasacchi, Marc Pollefeys, Thomas Funkhouser, et~al.
\newblock Openscene: 3d scene understanding with open vocabularies.
\newblock In \emph{Proceedings of the IEEE/CVF Conference on Computer Vision and Pattern Recognition}, 2023.

\bibitem[Qi et~al.(2019)Qi, Litany, He, and Guibas]{votenet}
Charles~R Qi, Or Litany, Kaiming He, and Leonidas~J Guibas.
\newblock Deep hough voting for 3d object detection in point clouds.
\newblock In \emph{proceedings of the IEEE/CVF International Conference on Computer Vision}, 2019.

\bibitem[Qi et~al.(2020)Qi, Chen, Litany, and Guibas]{imvotenet}
Charles~R Qi, Xinlei Chen, Or Litany, and Leonidas~J Guibas.
\newblock Imvotenet: Boosting 3d object detection in point clouds with image votes.
\newblock In \emph{Proceedings of the IEEE/CVF conference on computer vision and pattern recognition}, 2020.

\bibitem[Ramakrishnan et~al.(2021)Ramakrishnan, Gokaslan, Wijmans, Maksymets, Clegg, Turner, Undersander, Galuba, Westbury, Chang, et~al.]{hm3d}
Santhosh~K Ramakrishnan, Aaron Gokaslan, Erik Wijmans, Oleksandr Maksymets, Alex Clegg, John Turner, Eric Undersander, Wojciech Galuba, Andrew Westbury, Angel~X Chang, et~al.
\newblock Habitat-matterport 3d dataset (hm3d): 1000 large-scale 3d environments for embodied ai.
\newblock \emph{arXiv preprint arXiv:2109.08238}, 2021.

\bibitem[Reimers and Gurevych(2019)]{reimers-2019-sentence-bert}
Nils Reimers and Iryna Gurevych.
\newblock Sentence-bert: Sentence embeddings using siamese bert-networks.
\newblock In \emph{Proceedings of the 2019 Conference on Empirical Methods in Natural Language Processing}, 2019.

\bibitem[Roberts et~al.(2021)Roberts, Ramapuram, Ranjan, Kumar, Bautista, Paczan, Webb, and Susskind]{hypersim}
Mike Roberts, Jason Ramapuram, Anurag Ranjan, Atulit Kumar, Miguel~Angel Bautista, Nathan Paczan, Russ Webb, and Joshua~M Susskind.
\newblock Hypersim: A photorealistic synthetic dataset for holistic indoor scene understanding.
\newblock In \emph{Proceedings of the IEEE/CVF international conference on computer vision}, 2021.

\bibitem[Rozenberszki et~al.(2022)Rozenberszki, Litany, and Dai]{scannet200}
David Rozenberszki, Or Litany, and Angela Dai.
\newblock Language-grounded indoor 3d semantic segmentation in the wild.
\newblock In \emph{Proceedings of the European Conference on Computer Vision ({ECCV})}, 2022.

\bibitem[Rukhovich et~al.(2022{\natexlab{a}})Rukhovich, Vorontsova, and Konushin]{fcaf3d}
Danila Rukhovich, Anna Vorontsova, and Anton Konushin.
\newblock Fcaf3d: Fully convolutional anchor-free 3d object detection.
\newblock In \emph{European Conference on Computer Vision}, 2022{\natexlab{a}}.

\bibitem[Rukhovich et~al.(2022{\natexlab{b}})Rukhovich, Vorontsova, and Konushin]{imvoxelnet}
Danila Rukhovich, Anna Vorontsova, and Anton Konushin.
\newblock Imvoxelnet: Image to voxels projection for monocular and multi-view general-purpose 3d object detection.
\newblock In \emph{WACV}, pages 2397--2406, 2022{\natexlab{b}}.

\bibitem[Shi et~al.(2019)Shi, Wang, and Li]{PointRCNN}
Shaoshuai Shi, Xiaogang Wang, and Hongsheng Li.
\newblock Pointrcnn: 3d object proposal generation and detection from point cloud.
\newblock In \emph{IEEE Conference on Computer Vision and Pattern Recognition}, 2019.

\bibitem[Sima et~al.(2023)Sima, Tong, Wang, Chen, Wu, Deng, Gu, Lu, Luo, Lin, and Li]{occnet}
Chonghao Sima, Wenwen Tong, Tai Wang, Li Chen, Silei Wu, Hanming Deng, Yi Gu, Lewei Lu, Ping Luo, Dahua Lin, and Hongyang Li.
\newblock Scene as occupancy.
\newblock In \emph{Proceedings of the IEEE/CVF International Conference on Computer Vision}, 2023.

\bibitem[Simonelli et~al.(2019)Simonelli, Bulò, Porzi, López-Antequera, and Kontschieder]{MonoDIS}
Andrea Simonelli, Samuel Rota~Rota Bulò, Lorenzo Porzi, Manuel López-Antequera, and Peter Kontschieder.
\newblock Disentangling monocular 3d object detection.
\newblock In \emph{IEEE International Conference on Computer Vision}, 2019.

\bibitem[Song et~al.(2015)Song, Lichtenberg, and Xiao]{sunrgbd}
Shuran Song, Samuel~P Lichtenberg, and Jianxiong Xiao.
\newblock Sun rgb-d: A rgb-d scene understanding benchmark suite.
\newblock In \emph{Proceedings of the IEEE Conference on Computer Vision and Pattern Recognition}, 2015.

\bibitem[Straub et~al.(2019)Straub, Whelan, Ma, Chen, Wijmans, Green, Engel, Mur-Artal, Ren, Verma, et~al.]{replica}
Julian Straub, Thomas Whelan, Lingni Ma, Yufan Chen, Erik Wijmans, Simon Green, Jakob~J Engel, Raul Mur-Artal, Carl Ren, Shobhit Verma, et~al.
\newblock The replica dataset: A digital replica of indoor spaces.
\newblock \emph{arXiv preprint arXiv:1906.05797}, 2019.

\bibitem[Sun et~al.(2020)Sun, Kretzschmar, Dotiwalla, Chouard, Patnaik, Tsui, Guo, Zhou, Chai, Caine, et~al.]{waymo}
Pei Sun, Henrik Kretzschmar, Xerxes Dotiwalla, Aurelien Chouard, Vijaysai Patnaik, Paul Tsui, James Guo, Yin Zhou, Yuning Chai, Benjamin Caine, et~al.
\newblock Scalability in perception for autonomous driving: Waymo open dataset.
\newblock In \emph{Proceedings of the IEEE/CVF conference on computer vision and pattern recognition}, 2020.

\bibitem[Takmaz et~al.(2023)Takmaz, Fedele, Sumner, Pollefeys, Tombari, and Engelmann]{openmask3d}
Ay{\c{c}}a Takmaz, Elisabetta Fedele, Robert~W Sumner, Marc Pollefeys, Federico Tombari, and Francis Engelmann.
\newblock Openmask3d: Open-vocabulary 3d instance segmentation.
\newblock \emph{arXiv preprint arXiv:2306.13631}, 2023.

\bibitem[Tian et~al.(2023)Tian, Jiang, Yun, Wang, Wang, and Zhao]{occ3d}
Xiaoyu Tian, Tao Jiang, Longfei Yun, Yue Wang, Yilun Wang, and Hang Zhao.
\newblock Occ3d: A large-scale 3d occupancy prediction benchmark for autonomous driving.
\newblock \emph{arXiv preprint arXiv:2304.14365}, 2023.

\bibitem[Tian et~al.(2019)Tian, Shen, Chen, and He]{FCOS}
Zhi Tian, Chunhua Shen, Hao Chen, and Tong He.
\newblock Fcos: Fully convolutional one-stage object detection.
\newblock In \emph{IEEE Conference on Computer Vision and Pattern Recognition}, 2019.

\bibitem[Vora et~al.(2020)Vora, Lang, Helou, and Beijbom]{pointpainting}
Sourabh Vora, Alex~H Lang, Bassam Helou, and Oscar Beijbom.
\newblock Pointpainting: Sequential fusion for 3d object detection.
\newblock In \emph{Proceedings of the IEEE/CVF conference on computer vision and pattern recognition}, 2020.

\bibitem[Wald et~al.(2019)Wald, Avetisyan, Navab, Tombari, and Nie{\ss}ner]{3rscan}
Johanna Wald, Armen Avetisyan, Nassir Navab, Federico Tombari, and Matthias Nie{\ss}ner.
\newblock Rio: 3d object instance re-localization in changing indoor environments.
\newblock In \emph{Proceedings of the IEEE/CVF International Conference on Computer Vision}, 2019.

\bibitem[Wang et~al.(2021)Wang, Zhu, Pang, and Lin]{FCOS3D}
Tai Wang, Xinge Zhu, Jiangmiao Pang, and Dahua Lin.
\newblock {FCOS3D}: Fully convolutional one-stage monocular 3d object detection.
\newblock In \emph{Proceedings of the IEEE/CVF International Conference on Computer Vision (ICCV) Workshops}, 2021.

\bibitem[Wang et~al.(2022{\natexlab{a}})Wang, Pang, and Lin]{dfm}
Tai Wang, Jiangmiao Pang, and Dahua Lin.
\newblock Monocular 3d object detection with depth from motion.
\newblock In \emph{European Conference on Computer Vision (ECCV)}, 2022{\natexlab{a}}.

\bibitem[Wang et~al.(2022{\natexlab{b}})Wang, Xinge, Pang, and Lin]{pgd}
Tai Wang, ZHU Xinge, Jiangmiao Pang, and Dahua Lin.
\newblock Probabilistic and geometric depth: Detecting objects in perspective.
\newblock In \emph{Conference on Robot Learning}, pages 1475--1485. PMLR, 2022{\natexlab{b}}.

\bibitem[Wei et~al.(2023)Wei, Zhao, Zheng, Zhu, Zhou, and Lu]{surroundocc}
Yi Wei, Linqing Zhao, Wenzhao Zheng, Zheng Zhu, Jie Zhou, and Jiwen Lu.
\newblock Surroundocc: Multi-camera 3d occupancy prediction for autonomous driving.
\newblock In \emph{Proceedings of the IEEE/CVF International Conference on Computer Vision}, 2023.

\bibitem[Yadav et~al.(2023)Yadav, Ramrakhya, Ramakrishnan, Gervet, Turner, Gokaslan, Maestre, Chang, Batra, Savva, et~al.]{hm3d-semantic}
Karmesh Yadav, Ram Ramrakhya, Santhosh~Kumar Ramakrishnan, Theo Gervet, John Turner, Aaron Gokaslan, Noah Maestre, Angel~Xuan Chang, Dhruv Batra, Manolis Savva, et~al.
\newblock Habitat-matterport 3d semantics dataset.
\newblock In \emph{Proceedings of the IEEE/CVF Conference on Computer Vision and Pattern Recognition}, 2023.

\bibitem[Yan et~al.(2018)Yan, Mao, and Li]{SECOND}
Yan Yan, Yuxing Mao, and Bo Li.
\newblock Second: Sparsely embedded convolutional detection.
\newblock \emph{Sensors}, 18\penalty0 (10), 2018.

\bibitem[Zhang et~al.(2020)Zhang, Wang, and Loy]{moca}
Wenwei Zhang, Zhe Wang, and Chen~Change Loy.
\newblock Exploring data augmentation for multi-modality 3d object detection.
\newblock \emph{arXiv preprint arXiv:2012.12741}, 2020.

\bibitem[Zhou and Tuzel(2018)]{VoxelNet}
Yin Zhou and Oncel Tuzel.
\newblock Voxelnet: End-to-end learning for point cloud based 3d object detection.
\newblock In \emph{IEEE Conference on Computer Vision and Pattern Recognition}, 2018.

\bibitem[Zhou et~al.(2019)Zhou, Barnes, Lu, Yang, and Li]{6drot}
Yi Zhou, Connelly Barnes, Jingwan Lu, Jimei Yang, and Hao Li.
\newblock On the continuity of rotation representations in neural networks.
\newblock In \emph{Proceedings of the IEEE/CVF Conference on Computer Vision and Pattern Recognition}, 2019.

\bibitem[Zhu et~al.(2023{\natexlab{a}})Zhu, Zhang, Wang, Liu, and Chen]{object2scene}
Chenming Zhu, Wenwei Zhang, Tai Wang, Xihui Liu, and Kai Chen.
\newblock Object2scene: Putting objects in context for open-vocabulary 3d detection.
\newblock \emph{arXiv preprint arXiv:2309.09456}, 2023{\natexlab{a}}.

\bibitem[Zhu et~al.(2023{\natexlab{b}})Zhu, Ma, Chen, Deng, Huang, and Li]{3d-vista}
Ziyu Zhu, Xiaojian Ma, Yixin Chen, Zhidong Deng, Siyuan Huang, and Qing Li.
\newblock 3d-vista: Pre-trained transformer for 3d vision and text alignment.
\newblock In \emph{Proceedings of the IEEE/CVF International Conference on Computer Vision}, 2023{\natexlab{b}}.

\end{thebibliography}
}


\end{document}